\pdfoutput=1

\documentclass[11pt]{article}

\usepackage{acl}

\usepackage{times}
\usepackage{latexsym}
\usepackage[T1]{fontenc}
\usepackage[utf8]{inputenc}
\usepackage{microtype}

\usepackage{amsmath}
\usepackage{amssymb}
\usepackage{graphicx}
\usepackage{subcaption}          %
\usepackage{booktabs}
\usepackage{multirow}
\usepackage{enumitem}            %
\usepackage{xcolor}
\usepackage{mdframed} %
\usepackage[normalem]{ulem}      %
\usepackage{inconsolata}
\usepackage[disable]{todonotes}           %
\setuptodonotes{inline}          %

\graphicspath{{assets/}}

\newcommand{\onset}{t^\star}                                  %
\newcommand{\rr}{\mathrm{RR}}                                 %

\mdfdefinestyle{predictionbox}{
  linewidth=0.8pt, nobreak=true, innertopmargin=8pt, innerbottommargin=8pt,
  innerleftmargin=8pt, innerrightmargin=8pt, skipabove=8pt, skipbelow=10pt,
  backgroundcolor=black!3
}
\newmdtheoremenv[style=predictionbox]{prediction}{Prediction}

\title{Reliability Scales Inversely: Hallucinations Snowball Faster\\in Bigger Language Models}
\author{Kushal Chakrabarti \\
  Obviously Wrong, LLC \\
  San Francisco, CA 94107 \\
\texttt{kushalc@obviouslywrong.org}}

\begin{document}
\maketitle

\begin{abstract}
  Bigger language models are less reliable. Across three families, three benchmarks and six rungs,
  including in-the-wild chat logs, scaling closes the start-of-response knowledge gap up to $7\times$ while
  within-response knowledge \emph{degradation} grows up to $39\times$. We trace that residual to one variable, the
  per-position disagreement $\delta = \log p_M - \log p_O$ against a stronger oracle, whose second moment
  splits exactly into \emph{bias}$^2$ $\mathrm{KL}(p_M \,\|\, p_O)^2$ and \emph{decoding risk}
  $\mathrm{Var}[\delta]$. That split is an interpretability statement before it is a statistical one: the
  model's self-readable uncertainty $H(p_M)$ enters only the bias term, so the risk term has no
  model-readable component. Risk also takes a growing share of the squared error with scale, $31\%$ to
  $49\%$ from $1.7$B to $14$B. At a fabrication $H(p_M)$ relaxes within one token while risk persists up to
  $23\times$ longer, leaving a confident-but-precarious regime that bridges consecutive fabrications ($+69\%$ at
  $14$B). Contracting that risk at fixed $\mathrm{KL}$ removes $35$--$74\%$ of web-verified hallucinations
  across six rungs and three families. Semantic entropy fires $\approx$$30\%$ less on that branch
  ($p\!<\!10^{-16}$) though it carries nearly $4\times$ the fabrications. Bigger models snowball mistakes
  faster, through a failure mode that is dominant, self-perpetuating, causal and invisible to the model
  itself.
\end{abstract}

\begin{figure}[t!]
  \centering
  \begin{subfigure}{\columnwidth}
    \centering
    \includegraphics[width=0.96\linewidth]{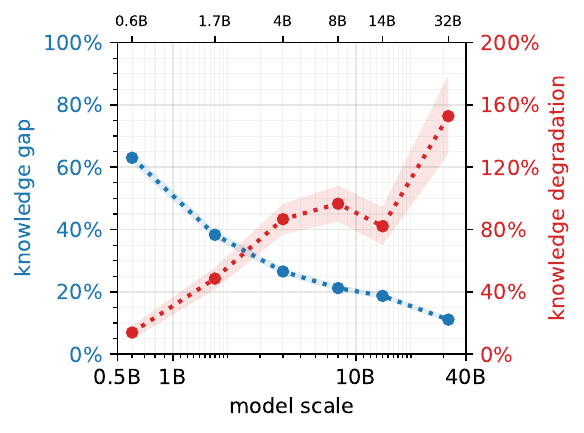}
    \phantomsubcaption  %
    \label{fig:indecision-a}
  \end{subfigure}

  \vspace{0.5em}

  \begin{subfigure}{\columnwidth}
    \centering
    \includegraphics[width=0.96\linewidth]{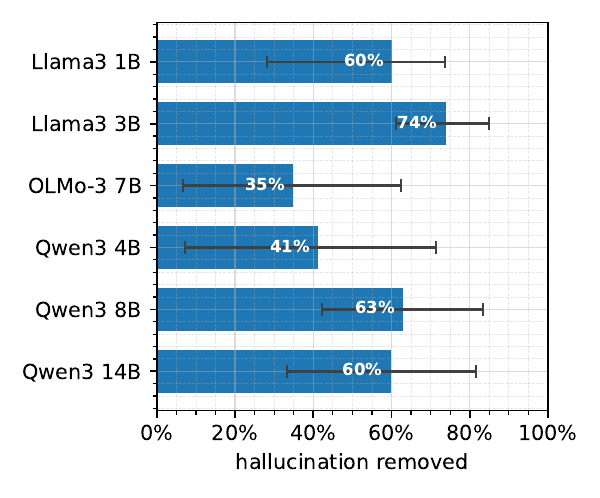}
    \phantomsubcaption  %
    \label{fig:indecision-b}
  \end{subfigure}

  \caption{
    \textbf{Reliability scales inversely; long-form hallucination is compounding risk.}
    \textbf{(a)} On LongFact++ free-run claims across the Qwen3 family ($0.6$--$32$B), the start-of-response
    knowledge gap (left) decreases while knowledge degradation over the full response (right)
    trends upward, so initial answers become truer but autoregressively decay faster as models scale
    (\autoref{sec:scaling:s1}).
    \textbf{(b)} A KL-preserving risk contraction (bias held fixed, drift $\ll|\mathrm{KL}|$)
    removes $35$--$74\%$ of web-verified hallucination across six models in three families, every $95\%$
    bootstrap CI excluding zero (\autoref{sec:mech:causal}).
    Capability and reliability are distinct scaling axes.
  }
  \label{fig:indecision}
\end{figure}

\section{Introduction}
\label{sec:intro}

The canonical account treats hallucination as a training-time knowledge gap \citep{kalai2026evaluating}, yet
three facts escape it: models fabricate around known facts \citep{simhi2025trust}; each fabrication makes the
next likelier within a single decode \citep{zhang2023snowball}; and decorrelating parallel representations
cuts hallucination at \emph{fixed} data and parameters \citep{chakrabarti2025neuraldiversity}. All three
implicate novel dynamics that causally mediate hallucination in modern language models.

We hypothesize that hallucination is a guess the model commits to, seeding an auto-regressive risk it can no longer
see. To analyze this, we decompose error into \emph{oracle surprise} (an oracle's cross-entropy on model
predictions), \emph{entropy} (the model's uncertainty), and \emph{risk} (the spread of their disagreement
across candidate tokens). Only entropy is self-readable, since a model feels its own uncertainty but neither the
oracle's surprise nor its risk.

\begin{figure*}[t!]
  \centering
  \begin{subfigure}{\textwidth}
    \centering
    \includegraphics[width=\textwidth]{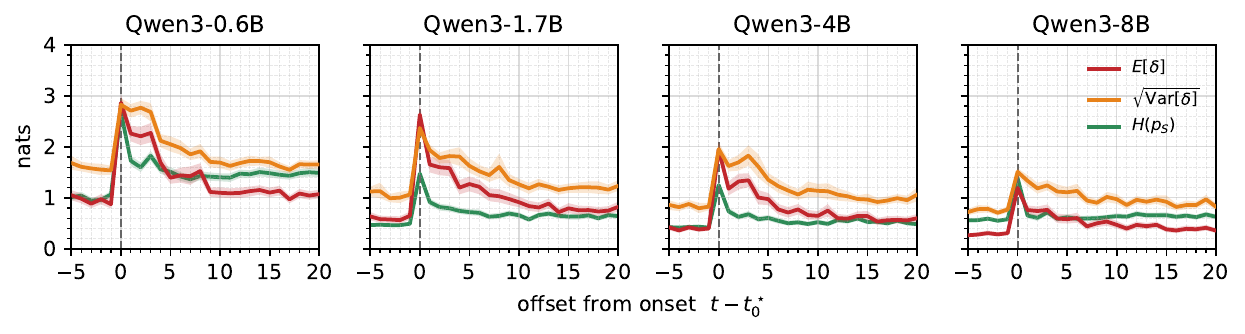}%
    \phantomsubcaption
    \label{fig:onset-norm}
  \end{subfigure}

  \par\vspace{0.2em}
  \begin{subfigure}{\textwidth}
    \centering
    \includegraphics[width=\textwidth]{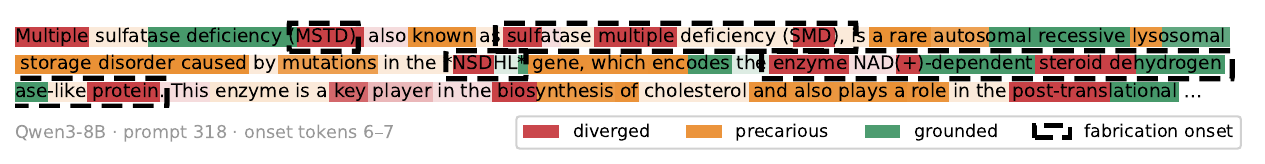}
    \phantomsubcaption
    \label{fig:token-risk}
  \end{subfigure}
  \caption{
    \textbf{At fabrication onset every channel spikes, but the model's self-readable uncertainty collapses
    within a token while the oracle-referenced decoding risk self-perpetuates.}
    \textbf{(a)} Onset-aligned free-run LongFact++ trajectories (per Qwen3 rung,
    $0.6$--$8$B vs.\ Qwen3-14B oracle; $95\%$ CIs): the bias gap $\mathrm{KL}(p_M\Vert p_O)$ (red),
    entropy $H(p_M)$ (green) and decoding risk $\sqrt{\mathrm{Var}[\delta]}$ (yellow) spike at the first
    unsupported claim ($t{=}t^\star_0$), but the self-readable entropy halves within a token while the
    oracle-referenced bias gap and risk linger downstream.
    \textbf{(b)} One illustrative Qwen3-8B trajectory, tokens colored by their model--oracle \emph{gap
    regime} (not a correctness label) over bias $\mathbb{E}[\delta] = \mathrm{KL}(p_M \,\|\, p_O)$,
    entropy~$H(p_M)$, risk $\sqrt{\mathrm{Var}[\delta]}$: \emph{grounded} (green; gap small in mean and
    spread), \emph{precarious} (amber; mean gap \& entropy low but risk high), \emph{diverged} (red; all
    high). It runs grounded, shifts at onset and disproportionately holds precarious downstream; the
    population version is panel (a).
  }
  \label{fig:onset}
\end{figure*}

Autoregression converts that risk into bias. At fabrication onset the model draws an
unsupportable token and conditions on it, freezing momentary risk into fixed bias (\autoref{sec:framework});
felt uncertainty relaxes while oracle-referenced risk persists (\autoref{sec:scaling:proxy}), and
fabrications chain, raising the next claim's hazard $1.08\times$--$1.71\times$ with scale
(\autoref{sec:mech:snowball}).

Between consecutive fabrications, the model disproportionately occupies a confident yet precarious regime
(low uncertainty, high risk), up to $+69\%$ (\autoref{sec:mech:bridge}). Contracting the risk
\emph{after} onset drops rest-of-response fabrication up to 74\% across three model families. This reframes
detectors that read functionals of $p_M$
\citep{kuhn2023semantic,farquhar2024detecting,manakul2023selfcheckgpt,chen2024inside}, which are sharp at
onset but blind to the subsequent risk driving downstream fabrications.

We demonstrate our hypothesis in four ways:
\begin{itemize}[leftmargin=1.3em, itemsep=2pt, topsep=2pt, parsep=0pt]
  \item \textbf{Hallucination Mechanism.} The moments of $\delta = \log p_M - \log p_O$ explain fabrications,
    and autoregression converts post-onset risk into hallucinatory bias (\autoref{sec:framework}).
  \item \textbf{Reliability Anti-Scaling.} Scaling closes the knowledge gap up to $7\times$ yet grows knowledge
    degradation up to $39\times$ across three benchmarks, including in-the-wild chat logs; capability
    and reliability are distinct scaling axes (\autoref{sec:scaling}).
  \item \textbf{Mechanistic Causality.} An over-confident risk regime bridges adjacent fabrications;
    contracting the risk at fixed KL prevents up to $74\%$ of web-verified hallucinations (\autoref{sec:mechanism}).
  \item \textbf{Detector Blindness.} That same regime structurally evades self-monitoring, leaving $p_M$-only
    detectors blind to the fabrications it drives (\autoref{sec:mech:blindness}).
\end{itemize}

\section{Self-Conditioning Converts Persistent Risk into Bias}
\label{sec:framework}

\begin{table*}[t]
  \centering
  \footnotesize
  \setlength{\tabcolsep}{2.5pt}
  \resizebox{\textwidth}{!}{%
  \begin{tabular}{l rrr r@{\,$\pm$\,}l r@{\,$\pm$\,}l r@{\,$\pm$\,}l r@{\,$\pm$\,}l r@{\,$\pm$\,}l r@{\,$\pm$\,}l }
    \toprule
    & \multicolumn{3}{c}{\textbf{support}} & \multicolumn{4}{c}{\textbf{FActScore}} & \multicolumn{4}{c}{\textbf{LongFact++}} & \multicolumn{4}{c}{\textbf{WildHallucinations}} \\
    \cmidrule(lr){2-4} \cmidrule(lr){5-8} \cmidrule(lr){9-12} \cmidrule(lr){13-16}
    \textbf{Model} & \textbf{FActS.} & \textbf{LF++} & \textbf{WildH.} & \multicolumn{2}{c}{\textbf{initial gap}} & \multicolumn{2}{c}{\textbf{degradation}} & \multicolumn{2}{c}{\textbf{initial gap}} & \multicolumn{2}{c}{\textbf{degradation}} & \multicolumn{2}{c}{\textbf{initial gap}} & \multicolumn{2}{c}{\textbf{degradation}} \\
    \midrule
    Qwen3-0.6B & 8.7\% & 37.1\% & 16.5\% & $97.0\%$ & ${\scriptstyle 1.6}$ & $\mathbf{1.6\%}$ & ${\scriptstyle 1.8}$ & $63.0\%$ & ${\scriptstyle 1.9}$ & $\mathbf{13.9\%}$ & ${\scriptstyle 4.4}$ & $89.0\%$ & ${\scriptstyle 3.5}$ & $\mathbf{3.3\%}$ & ${\scriptstyle 5.6}$ \\
    Qwen3-1.7B & 13.4\% & 51.7\% & 27.4\% & $90.1\%$ & ${\scriptstyle 2.8}$ & $5.8\%$ & ${\scriptstyle 3.4}$ & $38.3\%$ & ${\scriptstyle 1.5}$ & $48.5\%$ & ${\scriptstyle 6.9}$ & $65.6\%$ & ${\scriptstyle 5.2}$ & $34.0\%$ & ${\scriptstyle 9.7}$ \\
    Qwen3-4B & 28.1\% & 59.3\% & 33.1\% & $66.5\%$ & ${\scriptstyle 5.5}$ & $38.7\%$ & ${\scriptstyle 10.2}$ & $26.5\%$ & ${\scriptstyle 1.2}$ & $86.6\%$ & ${\scriptstyle 10.0}$ & $58.8\%$ & ${\scriptstyle 5.2}$ & $40.0\%$ & ${\scriptstyle 11.8}$ \\
    Qwen3-8B & 47.5\% & 65.0\% & 35.8\% & $42.4\%$ & ${\scriptstyle 6.1}$ & $63.2\%$ & ${\scriptstyle 23.6}$ & $21.2\%$ & ${\scriptstyle 1.0}$ & $96.6\%$ & ${\scriptstyle 11.6}$ & $49.8\%$ & ${\scriptstyle 5.6}$ & $74.8\%$ & ${\scriptstyle 17.3}$ \\
    Qwen3-14B & \textbf{67.2\%} & 68.7\% & \textbf{42.3\%} & $15.8\%$ & ${\scriptstyle 3.4}$ & $127.2\%$ & ${\scriptstyle 60.0}$ & $18.7\%$ & ${\scriptstyle 1.0}$ & $82.1\%$ & ${\scriptstyle 12.3}$ & $53.6\%$ & ${\scriptstyle 5.2}$ & $31.4\%$ & ${\scriptstyle 14.8}$ \\
    Qwen3-32B & 65.1\% & \textbf{75.2\%} & 37.1\% & $\mathbf{13.8\%}$ & ${\scriptstyle 3.1}$ & $229.6\%$ & ${\scriptstyle 79.4}$ & $\mathbf{11.1\%}$ & ${\scriptstyle 0.9}$ & $152.8\%$ & ${\scriptstyle 25.3}$ & $\mathbf{45.0\%}$ & ${\scriptstyle 4.8}$ & $89.7\%$ & ${\scriptstyle 18.0}$ \\
    \midrule
    OLMo-3-7B-Instruct & 43.2\% & 58.7\% & 28.4\% & $44.1\%$ & ${\scriptstyle 6.6}$ & $62.9\%$ & ${\scriptstyle 24.4}$ & $31.9\%$ & ${\scriptstyle 2.0}$ & $\mathbf{39.7\%}$ & ${\scriptstyle 11.6}$ & $62.7\%$ & ${\scriptstyle 6.6}$ & $\mathbf{44.5\%}$ & ${\scriptstyle 13.5}$ \\
    OLMo-3.1-32B-Instruct & \textbf{57.8\%} & \textbf{73.0\%} & \textbf{39.9\%} & $\mathbf{28.3\%}$ & ${\scriptstyle 6.2}$ & $\mathbf{24.5\%}$ & ${\scriptstyle 40.6}$ & $\mathbf{13.7\%}$ & ${\scriptstyle 1.0}$ & $136.7\%$ & ${\scriptstyle 21.4}$ & $\mathbf{51.2\%}$ & ${\scriptstyle 5.1}$ & $49.5\%$ & ${\scriptstyle 14.8}$ \\
    \midrule
    Llama-3.2-1B-Instruct & 30.9\% & 44.5\% & 21.1\% & $69.4\%$ & ${\scriptstyle 5.9}$ & $\mathbf{35.1\%}$ & ${\scriptstyle 10.1}$ & $50.9\%$ & ${\scriptstyle 1.8}$ & $\mathbf{25.7\%}$ & ${\scriptstyle 5.3}$ & $76.0\%$ & ${\scriptstyle 6.8}$ & $25.7\%$ & ${\scriptstyle 10.2}$ \\
    Llama-3.2-3B-Instruct & 53.6\% & 53.4\% & 29.6\% & $35.6\%$ & ${\scriptstyle 6.1}$ & $52.6\%$ & ${\scriptstyle 31.6}$ & $34.5\%$ & ${\scriptstyle 1.2}$ & $60.0\%$ & ${\scriptstyle 6.4}$ & $69.1\%$ & ${\scriptstyle 5.6}$ & $\mathbf{21.3\%}$ & ${\scriptstyle 10.3}$ \\
    Llama-3.1-8B-Instruct & \textbf{76.2\%} & \textbf{61.2\%} & \textbf{37.7\%} & $\mathbf{10.4\%}$ & ${\scriptstyle 2.4}$ & $174.2\%$ & ${\scriptstyle 79.0}$ & $\mathbf{25.5\%}$ & ${\scriptstyle 0.9}$ & $79.5\%$ & ${\scriptstyle 7.5}$ & $\mathbf{45.4\%}$ & ${\scriptstyle 5.8}$ & $91.5\%$ & ${\scriptstyle 21.7}$ \\
    \bottomrule
  \end{tabular}}
  \caption{\textbf{Scaling improves knowledge but worsens degradation: capability
    and reliability are bought separately.} Per eval and model, fact support vs.\ relative
    within-response position, both from a per-claim random-intercept logistic GLMM
    (\autoref{sec:scaling}, \autoref{tab:exp2-glmm}): \emph{initial gap} $=$ start-of-response
    hallucination rate; \emph{degradation} $=$ relative rise in hallucination from start to end ---
    the observable signature of the latent decoding noise; \emph{support} $=$ response-mean
    supported-fact rate. $\pm$ are $95\%$ intervals,
    best-in-family bold. Across families, tasks, and verifiers, the initial gap falls up to 7.0$\times$ while degradation
    rises up to 39$\times$ (excludes non-significant rungs).}
  \label{tab:scaling-twosided}
\end{table*}

Is hallucination just a knowledge gap? One disagreement variable says no: its moments recover cross-entropy
as gap and entropy as uncertainty, plus one they miss --- risk. Read mechanistically through autoregression
and training protocols, these terms yield four predictions on how fabrications arise, persist, propagate, and
evade self-monitoring at scale (\autoref{pred:dominance}--\ref{pred:blindness}).

\subsection{One variable's moments separate gap, uncertainty, and risk}
\label{sec:framework:moments}

At position $t$, with model $p_M(\cdot \mid y_{<t})$, oracle $p_O(\cdot \mid y_{<t})$, and a candidate token
$x$, define the \emph{disagreement variable} as the log-likelihood ratio
\begin{align}
  \delta_t(x) = \log p_M(x \mid y_{<t}) - \log p_O(x \mid y_{<t}).
\end{align}
We write $\delta$ for $\delta_t(V)$ at a random draw $V \sim p_M$ and $\delta(y_t)$ for
its realization on the sampled token; moments $\mathbb{E}[\cdot]$ and $\mathrm{Var}[\cdot]$ are taken over
$V \sim p_M$ unless noted, and every quantity we report is such a moment of $\delta$.

The first moment $\mathbb{E}[\delta]$ is the reverse $\mathrm{KL}$, and its second moment splits exactly around it:
\begin{align}
  \mathbb{E}[\delta^2]
  &= \underbrace{\mathrm{KL}(p_M \,\|\, p_O)^2}_{\text{bias}^2}
  + \underbrace{\mathrm{Var}[\delta]}_{\text{risk}}
  \label{eq:mse} \\
  &= [\underbrace{H(p_M, p_O)}_{\substack{\text{oracle}\\\text{surprise}}}
  - \underbrace{H(p_M)}_{\substack{\text{felt}\\\text{uncertainty}}}]^2 +
  \underbrace{\mathrm{Var}[\delta]}_{\text{risk}},
\end{align}
with the second line following by the standard identity $\mathrm{KL}(p_M \,\|\, p_O) = H(p_M, p_O) - H(p_M)$.

This split is an interpretability statement before it is a statistical one. The \emph{risk} term
$\mathrm{Var}[\delta]$, which depends on $p_O$ at every candidate the model entertains, has no
model-readable component. The model's \emph{felt uncertainty}
$H(p_M)$, the one self-readable functional that proxies risk, enters the error only inside the bias term, as
the subjective discount on \emph{oracle surprise}.

\subsection{Risk increasingly dominates error}
\label{sec:framework:dominance}

Although \autoref{eq:mse} is an algebraic identity, its two terms face different training pressures. Thus, which
dominates is an empirical question.

Maximum-likelihood pretraining \citep{olmo3} pressures the forward objective $H(p_{\text{data}}, p_M)$,
and both models approaching $p_{\text{data}}$ closes our mean $\mathrm{KL} \to 0$. Nothing
comparably narrows spread.

\begin{prediction}[Risk Concentration]\label{pred:dominance}
  As models scale, risk takes a growing share of error
  $\frac{\mathrm{Var}[\delta]}{\mathbb{E}[\delta^2]}$ and concentrates into heavier tails.
\end{prediction}

\subsection{Autoregression converts risk into bias}
\label{sec:framework:conversion}

At onset, sampling collapses the lottery $p_M(\cdot \mid y_{<t})$ to one token and appends it, converting
that token's risk (disagreement spread over candidates) to bias (fixed gap) inherited by every later step.

\begin{prediction}[Autoregressive Conversion]\label{pred:conversion}\label{pred:causal}
  At onset $t=\onset$, the sampling draw $y_\onset \sim p_M$ converts risk
  $\mathrm{Var}[\delta]$ into bias $\mathrm{KL}(p_M \,\|\, p_O)$.
\end{prediction}

The two terms then decay differently. $H(p_M)$ depends only on local next-token predictability, which fluent
continuation restores within a few tokens (\autoref{sec:scaling:proxy}). In contrast, $\mathrm{Var}[\delta]$
depends on the model--oracle gap over a premise now fixed in the shared context, which every subsequent token
inherits \citep{zhang2023snowball}.

\begin{prediction}[Persistence Asymmetry]\label{pred:persistence}
  For $t > \onset$, felt uncertainty $H(p_M)$ relaxes quickly to baseline while risk
  $\mathrm{Var}[\delta]$ persists.
\end{prediction}

\subsection{Fabrications hide from self-monitoring}
\label{sec:framework:blindness}

The asymmetry has a corollary that reaches past the model. Detectors built on sampling disagreement,
predictive and semantic entropy \citep{kuhn2023semantic,farquhar2024detecting,manakul2023selfcheckgpt} are
functionals of $p_M$, so by \autoref{eq:mse} they are blind to the risk term. Their lone foothold $H(p_M)$ spikes
at onset, then relaxes while $\mathrm{Var}[\delta]$ stays high.

\begin{prediction}[Detector Blindness]\label{pred:blindness}
  A $p_M$-only detector cannot read risk $\mathrm{Var}[\delta]$ or effectively detect risk-mediated fabrications.
\end{prediction}

\autoref{sec:scaling} tests Risk Concentration and Persistence Asymmetry on the moments themselves;
\autoref{sec:mechanism} escalates Autoregressive Conversion and Detector Blindness from observation to a
causal contraction that isolates risk.

\begin{figure*}[t]
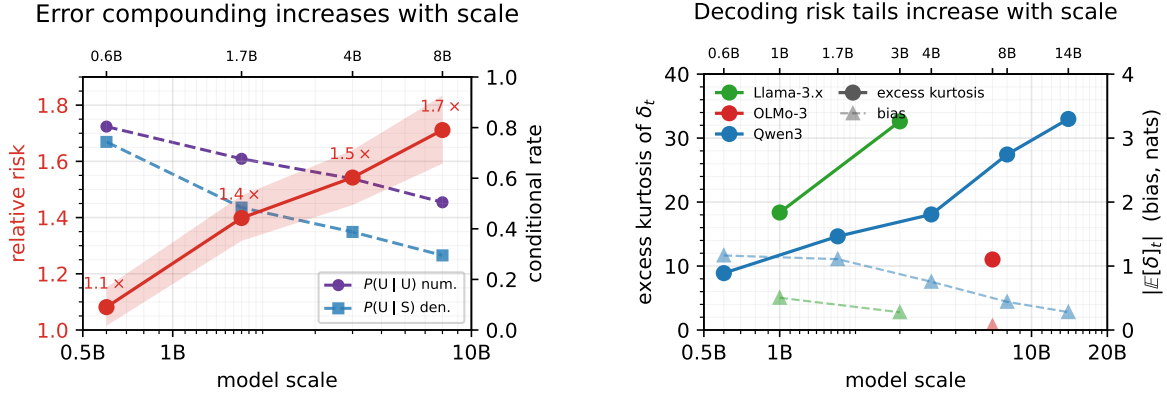

  \centering
  \input{assets/fig_exp8a_snowball_vs_scale_float.tex}
  \hfill
  \input{assets/fig_exp4c_within_position_kurtosis_float.tex}
  \caption{\textbf{The variance side of hallucination strengthens with scale in two ways.}
    \emph{(a)} the across-claim compounding (relative risk $\rr$ that a fabrication begets the next) rises
    monotonically with Qwen3 model size, and \emph{(b)} the per-position oracle-gap tail (excess kurtosis
    of $\delta_t$) sits far above the Gaussian baseline and rises monotonically while the mean gap shrinks in
    both Llama-3.2 and Qwen3 (OLMo-3 corroborating at $7$B).
  }
  \label{fig:scale-trends}
\end{figure*}

\section{Scaling Closes the Knowledge Gap and Grows the Risk}
\label{sec:scaling}

As models scale, where does the error go? We show that as oracle surprise reduces, it re-organizes into risk.
Reliability degrades faster than capability improves (\autoref{sec:scaling:s1}), the model's proxy stops
tracking risk after onset (\autoref{sec:scaling:proxy}), and risk grows with scale in both share and tail
(\autoref{sec:scaling:s3}).

\subsection{Reliability degrades faster than knowledge improves}
\label{sec:scaling:s1}

Reliability degrades several-fold faster than knowledge improves. Across families, tasks, and verifiers, the
initial knowledge gap closes up to $7.0\times$ while auto-regressive knowledge degradation grows up to $39\times$
(\autoref{tab:scaling-twosided}), estimated by a per-claim mixed-effects logistic fit
(\autoref{tab:exp2-glmm}; stat. sig. rungs). The trend extends beyond curated benchmarks: on entities drawn
from real user--chatbot conversations (\citealp{zhao2024wildhallucinations}), degradation remains significant
at $10$ of $11$ rungs across all three families (\autoref{app:scaling}). The error reflects not what the
model fails to \emph{know} but how it \emph{commits} among candidates it already ranks correctly --- the risk
term of \autoref{eq:mse}.

\subsection{Decoding breaks the model risk proxy}
\label{sec:scaling:proxy}

Felt uncertainty $H(p_M)$ is the only self-readable functional of $p_M$ that can proxy decoding risk, and
at the onset it tracks it: $\mathbb{E}[\delta]$, $H(p_M)$, and $\sqrt{\mathrm{Var}[\delta]}$ rise contemporaneously
(\autoref{fig:onset-norm}).

After onset, however, the channels diverge (\autoref{pred:persistence}). The self-readable $H(p_M)$ collapses fastest
(half-life $t_{1/2} <1$ tokens across Qwen3), while the oracle-referenced $\mathbb{E}[\delta]$ and
$\sqrt{\mathrm{Var}[\delta]}$ both require $3$--$5$ tokens to decay by 50\%. Once
appended, the sampled token fixes bias and risk while local fluency quickly recovers; no functional of
$p_M$ tracks persistent risk after onset.

These dynamics are robust across oracle scale (14B vs.\ 32B Qwen3), oracle family (Qwen3 vs.\ DeepSeek-V3), and
oracle inference settings (top-k $\in \{1024, 2048, 4096\}$, \autoref{app:oracle}).

\subsection{Risk worsens with scale}
\label{sec:scaling:s3}

\begin{figure*}[t]
  \centering
  \includegraphics[width=\textwidth]{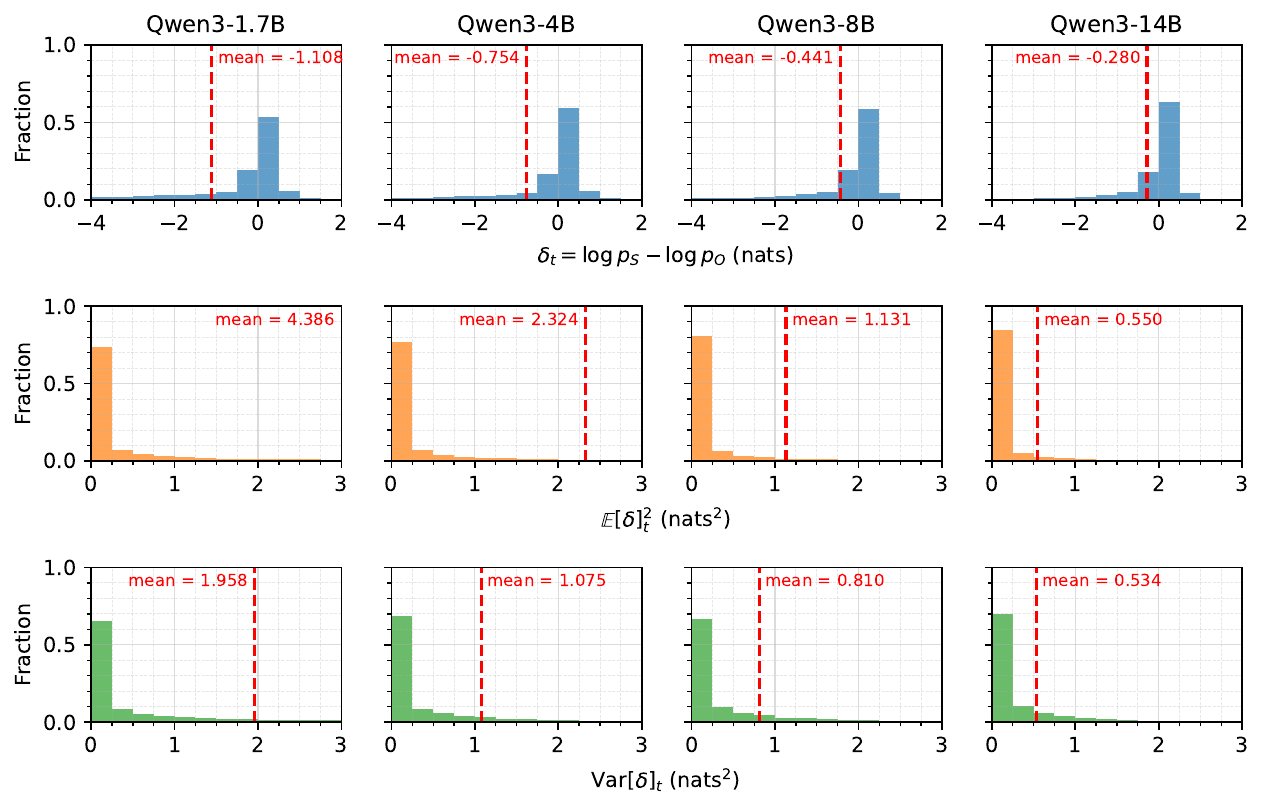}
  \caption{
    \textbf{Decoding risk is a growing share of the per-position error with scale.} Teacher-forced
    $\delta_t = \log p_M(y^*_t) - \log p_O(y^*_t)$ (Qwen3 vs.\ Qwen3-32B; $14{,}629$ tokens/model, $58$
    biographies; $0.6$B omitted). \emph{Top:} realized-token $\delta_t$ density, leptokurtic with scale.
    \emph{Middle/bottom:} the distributional bias$^2$ ($\mathrm{KL}(p_M\Vert p_O)_t^2$) and risk
    ($\mathrm{Var}_{p_M}[\delta]_t$) terms (nats$^2$); both shrink, but the bias$^2$
    term faster, so risk's share of $\mathbb{E}[\delta^2]_t$ rises $31\%\to49\%$ from $1.7$ to
    $14$B. Cross-sectional view of the onset-aligned breakdown in \autoref{fig:onset-norm}.
  }
  \label{fig:xsection}
\end{figure*}

Risk worsens with scale along two axes: its tail and its share of the error. Realized on the model's
outputs, the per-position risk $\mathrm{Var}[\delta]_t$ shrinks more slowly than the bias$^2$ term of
\autoref{eq:mse}, so its share of the squared error climbs $31\%$ to $49\%$ between $1.7$B and $14$B
(\autoref{fig:xsection}). Simultaneously, the gap grows heavier-tailed, with excess kurtosis of $\delta$ rising
$8.9\to33.0$ (\autoref{fig:tail}). Counterintuitively, a larger model agrees with the oracle more often on
average, yet diverges further when it does.

The heavy tail is not diffuse; it localizes to identifiable tokens. A three-state Markov-switching fit on
$\delta$ (\autoref{fig:token-risk}, \autoref{tab:exp8b-msar}) resolves decoding into \emph{grounded},
\emph{precarious}, and \emph{diverged} regimes. It assigns unsupported tokens to the correct state
$\approx$$1.7\times$ as often as supported ones (AUROC $0.68$--$0.71$), acting as a label-free white-box
marker of risk that holds across scale.

Both trends confirm Risk Concentration (\autoref{pred:dominance}); the onset dynamics of
\autoref{sec:scaling:proxy} bear out Persistence Asymmetry (\autoref{pred:persistence}).

\section{A Risk Regime Bridges Fabrications}
\label{sec:mechanism}

What does the lingering risk do? Observationally, a fabrication raises the next claim's hazard
(\autoref{sec:mech:snowball}) through a confident yet precarious risk regime on the inter-claim bridge
(\autoref{sec:mech:bridge}); causally, contracting that regime cuts downstream fabrications
(\autoref{sec:mech:causal}), which the model's own signals never register (\autoref{sec:mech:blindness}).

\subsection{Fabrications raise fabrication hazard}
\label{sec:mech:snowball}

A committed fabrication raises the next claim's hazard beyond the topic base rate. The relative
risk $\frac{P(U_{j+1}\mid U_j)}{P(U_{j+1}\mid S_j)}$ of an unsupported claim $U_{j+1}$ given an unsupported
vs.\ supported predecessor ($U_j$ vs.\ $S_j$) climbs monotonically $1.08\to1.71\times$ with scale, every
$95\%$ bootstrap CI excluding $1$ (\autoref{fig:snowball}).

To control topic difficulty we restrict to \emph{mixed} responses (both supported and unsupported claims);
the lift holds in both samples, every CI excluding $1$ over thousands of pairs (\autoref{tab:exp8a}).

The compounding deepens even as fabrications grow rarer. The spontaneous rate $P(U_{j+1}\mid S_j)$ falls
$0.74\to0.29$ with scale, so each committed fabrication is increasingly self-perpetuating.

\subsection{A risk regime sits between fabrications}
\label{sec:mech:bridge}

\begin{figure}[t]
  \centering
  \includegraphics[width=\columnwidth]{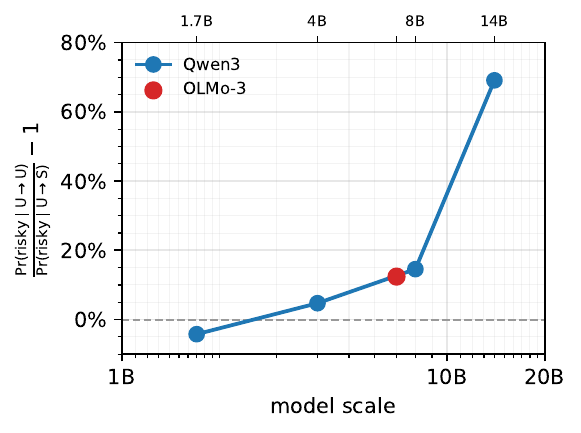}
  \caption{
    \textbf{An over-confident risk regime increasingly bridges adjacent fabrications.} The $y$-axis is the
    relative likelihood that a bridge token is in the \emph{precarious} regime
    (\autoref{fig:token-risk}) when one fabrication leads to another rather than to a supported claim:
    $\frac{P(\text{prec}\mid\text{U}{\to}\text{U})}{P(\text{prec}\mid\text{U}{\to}\text{S})}-1$. Each point is one
    model scale ($x$, log), over adjacent pairs (gap $\le 10$ tokens, sample $n \ge 100$) in mixed responses.
  }
  \label{fig:bridge}
\end{figure}

The \emph{precarious} regime (confident yet high-risk) is the bridge between two consecutive fabrications.
Re-scoring each free-run trajectory with the per-token (bias, entropy, risk) state model of \autoref{fig:token-risk}
(\autoref{app:lowcommit}), this regime disproportionately occupies the inter-claim bridge. Furthermore, the
excess likelihood $\frac{P(\text{prec}\mid U_j,U_{j+1})}{P(\text{prec}\mid U_j,S_{j+1})}-1$ of being in the
precarious regime climbs from near zero at the smaller rungs to $+15\%$ at $8$B and $+69\%$ at $14$B
(\autoref{fig:bridge}, adjacent pairs, gap $\le 10$ tokens, mixed responses). This is the observational
signature predicted by \autoref{pred:conversion}.

\subsection{Contracting that risk cuts fabrications}
\label{sec:mech:causal}

Removing decoding risk at fixed mean gap causally lowers downstream hallucination across three model
families --- an \emph{in silico} knockout of \autoref{sec:framework}'s conversion. An on-policy, co-resident oracle
scores the live support gap at every step; at each position crossing a per-model divergence threshold we
replace the model's next-token distribution $p_M$ with the mean-preserving contraction
\begin{equation}
  q_\lambda(x)\propto p_M(x)\,e^{-\lambda(\delta(x)-\mu)^2+\nu\,\delta(x)},
  \label{eq:contraction}
\end{equation}
where $\mu=\mathbb{E}_{p_M}[\delta]=\mathrm{KL}(p_M\Vert p_O)$ is the mean gap, $\nu$ is solved per step so that
$\mathbb{E}_{q_\lambda}[\delta]=\mu$ \emph{exactly}, and the dose
$\rho=\mathrm{Var}_{q_\lambda}[\delta]/\mathrm{Var}_{p_M}[\delta]<1$ sets the target variance fraction via
$\lambda$. The quadratic penalty down-weights tokens whose gap departs from $\mu$ while the $\nu\,\delta$ term
re-pins the mean, so \autoref{eq:contraction} is a pure variance knob at fixed $\mathrm{KL}$ (swept grid in
\autoref{app:contraction}), after which we free-run and verify every downstream claim.

\begin{figure*}[t]
  \centering
  \includegraphics[width=\textwidth]{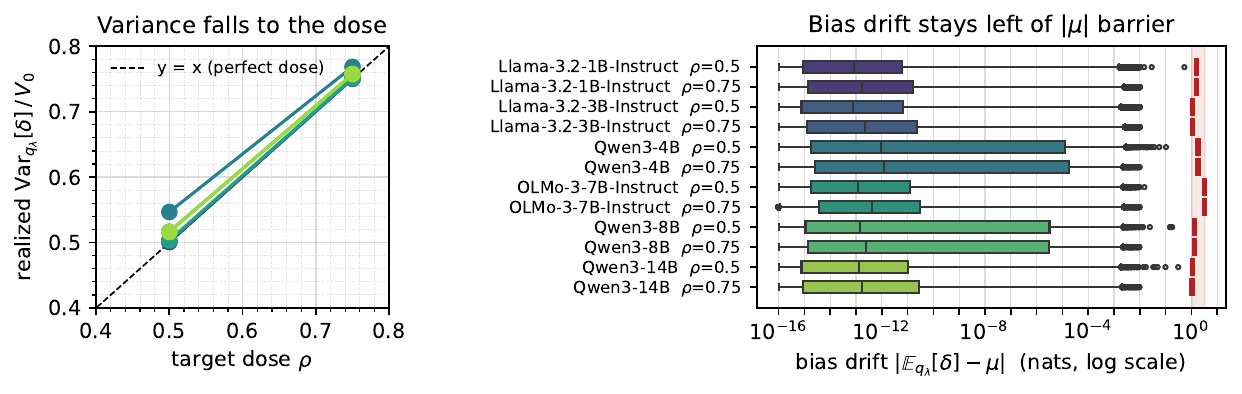}
  \caption{
    \textbf{Manipulation check: the contraction moves variance and nothing else.}
    Left: realized per-fired-step variance fraction $\mathrm{Var}_{q_\lambda}[\delta]/V_0$ vs target dose $\rho$,
    per rung --- points on the identity mean variance fell to the dose (realized $0.50$--$0.77$ at
    $\rho{=}0.5$--$0.75$).
    Right: bias drift $|\mathbb{E}_{q_\lambda}[\delta]-\mu|$ (box = IQR, whiskers = 5--95\%) against the $|\mu|$
    barrier (red) the contraction must not cross. Median drift is $\le 10^{-11}$ nats, $\sim$11 orders below
    $|\mu|$, demonstrating that the downstream effect (\autoref{tab:exp9c-causal}) traces to decoding risk,
    not a mean shift or topic difficulty.
  }
  \label{fig:exp9c-bias-hold}
\end{figure*}

The manipulation check (\autoref{fig:exp9c-bias-hold}) confirms realized variance falls to the dose $\rho$ while
per-step mean drift stays $\sim$11 orders of magnitude below the bias it must preserve, so any downstream
change is attributable to the decoding-risk channel, not the mean gap or topic difficulty.

Contracting risk cuts fabrication at every rung. Relative to control on the same trajectories, the best arm
lowers the rest-of-response unsupported-claim rate by a within-trajectory paired $\Delta$ of $0.13$--$0.33$,
a $35$--$74\%$ relative reduction across all six rungs and three families, with the $95\%$ bootstrap CI
excluding zero at every rung (\autoref{tab:exp9c-causal}, \autoref{fig:indecision-b}).

The bridge carries the effect. Contracting risk only in the inter-claim gap (after the last token
of an unsupported claim and before the first token of the next claim) prevents $21$--$60\%$ of
the second fabrications. It is significant at every rung across three families and grows with scale
(\autoref{tab:exp9d-bridge}).

Finally, we show confounds cannot carry the effect. Response length ($\pm0.5\%$), atomic-claim count
($\pm10\%$), and self-BLEU
diversity ($1.0\%$ mean) move slightly and non-directionally relative to control --- far too small to explain
a $35$--$74\%$ drop in unsupported fraction (\autoref{tab:exp9c-confounds}).

\subsection{Self-monitoring is blind to the bridge}
\label{sec:mech:blindness}
\begin{table}[t]
  \centering
  \small
  \setlength{\tabcolsep}{6pt}
  \begin{tabular}{l rrr rr}
    \toprule
    & \multicolumn{3}{c}{\textbf{SemEnt}} & \multicolumn{2}{c}{\textbf{count}} \\
    \cmidrule(lr){2-4} \cmidrule(lr){5-6}
    \textbf{Model} & onset & bridge & gap & onset & bridge \\
    \midrule
    Q0.6B & $0.365$ & $0.264$ & $0.101^{***}$ & $763$ & $4{,}588$ \\
    Q1.7B & $0.296$ & $0.204$ & $0.092^{***}$ & $737$ & $2{,}754$ \\
    Q4B & $0.306$ & $0.203$ & $0.102^{***}$ & $708$ & $2{,}110$ \\
    Q8B & $0.255$ & $0.167$ & $0.087^{***}$ & $650$ & $1{,}448$ \\
    \bottomrule
  \end{tabular}
  \caption{\textbf{Semantic entropy goes blind on the bridge.} Mean semantic entropy (higher $=$ more strongly flagged) on verified-\emph{unsupported} claims, by run structure: the \emph{onset} fabrication vs.\ the strictly-consecutive \emph{bridge} that follows. The $p_M$-only detector fires less on the bridge at every Qwen3 rung ($^{***}\,p<10^{-3}$, one-sided Mann--Whitney), despite it carrying most fabrications ($10{,}900$ vs.\ $2{,}858$ at onset).}
  \label{tab:detector-blindness}
\end{table}

The committed branch is also where the model's own signals weaken, bearing out Detector Blindness
(\autoref{pred:blindness}). We score each web verified-\emph{unsupported} claim with Semantic Entropy
(\citet{kuhn2023semantic,farquhar2024detecting}; \autoref{app:detectors}). At every Qwen3 rung it fires less
on the committed branch than at onset ($28$--$34\%$ lower, gap $0.09$--$0.10$ nats, Mann--Whitney $p <
10^{-16}$; \autoref{tab:detector-blindness}) despite holding nearly $4\times$ as many fabrications as onset.

Taken together, these tests converge on a single culprit: the low-bias, high-risk inter-claim regime, which
causally compounds downstream fabrications (establishing Autoregressive Conversion,
\autoref{pred:conversion}) yet stays hidden from self-monitoring (establishing Detector Blindness,
\autoref{pred:blindness}). The dominant failure mode is the invisible one.

\section{Related Work}
\label{sec:related}

Error accumulation in generation is not new. \citet{zhang2023snowball} name \emph{hallucination snowballing},
where a model commits early, then fluently defends what it could recognize as false in isolation; later work plants
hallucinatory context in vision-language models \citep{zhong2024mmhalsnowball} or ties faithfulness decay to
attention dynamics \citep{yang2025hallucinate}. Such decay is read as exposure bias \citep{arora2022exposure},
treated at training time by scheduled sampling \citep{bengio2015scheduled} and sequence-level objectives
\citep{ranzato2016sequence}. We give the first on-policy, causal, whole-response test on a frozen model, isolating
self-conditioning, a property of decoding not training, as the cause.

Whether that failure is a sampling phenomenon or a knowledge deficit organizes much of the literature
\citep{ji2023survey,huang2024survey}. Sampling-side evidence is direct, spanning architecture, activations, and
decoding. Decorrelating parallel representations
cuts hallucination at \emph{fixed} parameter and data budgets \citep{chakrabarti2025neuraldiversity}, injected
hidden-activation noise exposes the same variance \citep{liu2025noiseinjection}, and temperature and nucleus
truncation trade factuality for diversity \citep{lee2022factuality,holtzman2020curious}, tunable by
hallucination-aware thresholds \citep{chang2025real}. \citet{abbasiyadkori2024believe} separate epistemic from
aleatoric uncertainty by iterative prompting within one model. We take the trade-off as given and ask what
converts onset sampling variance into persistent error, measuring it with an \emph{inter}-model split against a
stronger oracle rather than within the model itself.

Wrong outputs are frequently not missing knowledge. \citet{gekhman2024finetuning} define a \emph{known fact} as
greedy-correct (our definition) and show fine-tuning on unknown facts breeds hallucination; \citet{orgad2025llmsknow}
find internal states encode the right answer the model contradicts; \citet{simhi2025trust} show hallucinations
strike known facts with high certainty, not sampling noise. Such self-knowledge is also steerable and
localizable. Entity latents gate
refusal versus hallucination \citep{ferrando2025entity}, calibration improves with scale \citep{kadavath2022language},
and non-factual recall localizes to layers \citep{yu2024mechanistic}. Our decomposition subsumes these as a small bias
with non-zero risk, the distributional, decoding-time form of their static, probe-level gap.

Methods already exploit that residual, since sample disagreement is itself a hallucination signal. Predictive
uncertainty was an early cue \citep{xiao2021hallucination}; semantic entropy clusters generations by meaning to
flag confabulations \citep{kuhn2023semantic,farquhar2024detecting}, SelfCheckGPT detects from black-box
disagreement \citep{manakul2023selfcheckgpt}, linear probes recover it from one generation \citep{kossen2024sep},
and hidden-state covariance scores hallucination directly \citep{chen2024inside}, as a survey catalogs
\citep{kang2025uncertainty}. These build detectors from our across-sample variance; we add the time-resolved
contrast they lack, showing that the self-readable side goes quiet (\autoref{fig:onset-norm}) while risk
persists, feeding downstream bias along the committed branch.

Orthogonally, theory-side accounts locate the cause in training rather than decoding. On these accounts,
calibrated models must hallucinate on rare facts \citep{kalai2024calibrated}, evaluations reward guessing over abstention
\citep{kalai2026evaluating}, dominant associations overshadow relevant knowledge \citep{zhang2025overshadowing},
and hallucinations emerge with factual knowledge during training \citep{zucchet2025facts}. These explain why
hallucination is inevitable at pretraining and under current incentives; none predicts a causal self-conditioning
effect that \emph{strengthens} with scale.

The nascent science of reliability casts it as an axis distinct from capability \citep{rabanser2026reliability},
with larger models growing more capable yet answering confidently wrong \citep{zhou2024larger} and autonomy
bounded by reliably-completed task length \citep{kwa2025horizon}. These measure it behaviorally and extrinsically, via
prompt-perturbation sensitivity and accuracy across a human-difficulty axis. We demonstrate a stronger, mechanistic
result inside the model: at a fixed prompt, unreliability intensifies within a response even as scale grows
capability, and we causally isolate its origin to self-conditioning, intrinsic to autoregressive decoding.

Our reliability result instantiates \emph{inverse scaling}, tasks on which larger models do worse
\citep{mckenzie2023inverse}. The closest precedents are factual, where larger models score lower on TruthfulQA by
imitating human misconceptions \citep{lin2022truthfulqa} and sycophancy strengthens with scale and RLHF
\citep{perez2023discovering}. Our degradation adds another, fundamental form that we trace back to a
structural aspect of auto-regressive language models: decoding-time self-conditioning.

\section{Discussion and Implications}
\label{sec:discussion}

The field treats hallucination today as a knowledge gap. Missing mass, inevitable for rare facts and rewarded
by current evaluations, can be bought down by scale
\citep{kalai2024calibrated,kalai2026evaluating,zhang2025overshadowing}.
Detectors catch the residue by reading the model's uncertainty
\citep{kuhn2023semantic,farquhar2024detecting,manakul2023selfcheckgpt,chen2024inside}. Both assume error
tracks what the model does not know.

Building on \citet{rabanser2026reliability}, we show that not only is reliability an independent axis, but it
scales \emph{inversely} with parameters at a fixed knowledge gap via decoding dynamics. Splitting
model--oracle disagreement into gap, uncertainty, and risk (\autoref{sec:framework}) separates what a model
does not know from how it \emph{decodes} what it knows. Scale buys knowledge while risk-driven degradation grows
(\autoref{sec:scaling}); contracting risk at fixed KL cuts downstream fabrication across three families
(\autoref{sec:mechanism}).

Two open problems follow, split by access and purpose. The detection challenge is black-box: because many
detectors read functionals of $p_M$ and \autoref{eq:mse} shows $\mathrm{Var}[\delta]$ is not one, they
underperform on unreliability (\autoref{sec:mech:blindness}). To effectively detect unreliability,
increasingly mediated by risk in modern language models, black-box detectors must cheaply carry an external referent.

The interpretability question is white-box: where does risk live and which components mediate it? Early evidence offers
clues: our gap-only fit reads risk without labels (\autoref{sec:scaling:s3}) and representation interventions
move hallucination at fixed budgets \citep{chakrabarti2025neuraldiversity,ferrando2025entity}, though none
point to a component. Localizing risk to layers, heads or features would expand our understanding of model
behavior under the second moment, which our work implicates as fundamental to reliability.

The failure mode that dominates the error budget, self-perpetuates across a response, and worsens with scale
is exactly the one self-monitoring structurally cannot capture. As such, reliability will not arrive for free
with capability.

\section*{Limitations}
\label{sec:limitations}

\paragraph{The causal test is a ceiling.}
The causal test (\autoref{sec:mech:causal}) uses a co-resident oracle to both detect onset and define the
contraction, so it bounds an online variance controller, not a deployable oracle-free rule. The oracle is what
isolates the knob: a mean-shifting one (temperature, self-grounding) cannot perturb $\mathrm{Var}[\delta]$ at
fixed $\mu$. It conditions on trajectories that reach an onset, so generalization beyond them is unestablished,
though the contrast stays fair --- control and intervention draw from one free-run conditional distribution. The effect is same-signed and significant
($95\%$ CI excludes zero) at all six model$\times$family rungs, though OLMo-3 (one rung) and Llama-3.2 (two
low-$N$ rungs) sample non-Qwen families thinly.

\paragraph{Detector comparison is mechanistic.}
We do not claim to out-detect named methods. The prediction is structural: every self-readable signal is a
functional of $p_M$ (\autoref{eq:mse}), and the one we resolve in time, $H(p_M)$, relaxes within a token of
onset while oracle-referenced risk persists (\autoref{fig:onset-norm}). Semantic entropy and SelfCheckGPT, also
$p_M$-functionals, inherit that relaxation; a time-resolved head-to-head on the inter-claim bridge is left open.

\paragraph{Decoding and position models.}
The moment results are teacher-forced --- computed from $p_M$ and $p_O$ directly, not from sampled tokens ---
hence invariant to the decoding temperature and top-$p$ a deployment would choose. Only the free-run analyses
(onset, snowball, intervention) depend on decoding, run at each model's standard nucleus setting; re-running the
snowball at greedy and $\tau{=}0.5$ keeps the across-claim relative risk above $1$ and scale-increasing
(\autoref{tab:decoding-ablation}), though a denser sweep would tighten it.
Knowledge degradation (\autoref{tab:scaling-twosided}) is a per-claim logistic GLMM (response-clustered
start-to-end rise in hallucination; \autoref{tab:exp2-glmm}), not a linear-probability slope$\div$intercept
ratio, so it avoids the wide intervals that ratio carries at a small start-of-response gap; the \emph{gap}
column remains an OLS intercept.

\paragraph{Domain scope.}
The evidence is English long-form parametric generation: FActScore biography prompts ($64$ topics) drive the position fit
and teacher-forced decompositions, LongFact++ open-domain prompts ($401$ topics) supply the free-run trajectories for the onset, snowball,
bridge, and contraction analyses (\autoref{app:setup}), and WildHallucinations extends the position fit to
in-the-wild entities ($20$ categories) judged against curated web corpora (\autoref{app:setup}). Reasoning chains,
source-grounded generation, summarization, dialogue, code, and other languages lie outside this paper's
scope; the moment analyses (\autoref{sec:framework}--\ref{sec:scaling}) are scoped to settings where
$\delta$ has a verifiable referent.

\paragraph{Verifier.}
Claim labels come from a Claude web-search verifier, so verifier error could propagate. Our guard is
independent corroboration: the verifier-free white-box risk marker agrees with the labels at AUROC $0.68$--$0.71$
across scale (\autoref{app:lowcommit}), and the effects replicate across three families and several metrics.
Early positions are the most verifier-sensitive; a human spot check, double-judge pass, and verifier-temperature
sweep are open.

\section*{Ethics Statement}

This work analyzes an existing failure mode and adds no generative capability. Its intended use is
protective: the risk signal (\autoref{sec:scaling:s3}) supports abstention, calibrated confidence, routing
risky spans to verification, and transparent disclosure of model reliability. The dual-use surface is that
any such signal could be inverted to push fabrications into the high-commitment regime where detectors go
quiet (\autoref{sec:discussion}); two things bound this: the signal needs white-box logit access, and
publishing the mechanism aids detection at least as much as evasion. We make no deployed-system, clinical, or
legal claims; all data are public benchmarks scored with a public verifier API, at modest single-GPU compute.

\section*{Use of Large Language Models}

LLMs were used as a compilation tool to assist with the writing and organizing sections of this paper,
including literature review synthesis, section structuring, LaTeX formatting, and co-generation of
experimental code. All technical content, experimental design, theoretical contributions, and scientific
claims are the author's original work.

\section*{Acknowledgements}

We thank Gabriel Nakajima An, Jimmy Jin, Wittawat Jitkrittum, Iman Modarressi, Joshua Penman, and Barak
Widawsky, whose feedback sharpened both the argument and the experiments behind it; Anthropic, for the API
credits that funded the web-search claim verification throughout; and the research community at South Park
Commons, for support and discussion at every stage of this work.

\bibliography{anthology,references}

\appendix
\onecolumn

\counterwithin{figure}{section}
\counterwithin{table}{section}

\section{Models, Data, and Verifier Protocol}
\label{sec:appendix}
\label{app:setup}

\paragraph{Models.}
Results span three open model families. The Qwen3 family \citep{qwen3technical} ($0.6$--$32$B;
\autoref{tab:models}) carries the scaling sweep, the cross-sectional and over-commitment analyses,
and the position fit; the OLMo-3 family \citep{olmo3} adds an independent second-family rung
(OLMo-3-7B-Instruct model, OLMo-3.1-32B-Instruct oracle; the family ships only these two sizes)
and the Llama-3 family \citep{llama3} a third (Llama-3.2-1B/3B-Instruct models, Llama-3.1-8B-Instruct
oracle) to the onset-dynamics, bridge, and intervention results. All run in their no-thinking configuration.
Holding architecture fixed across scale within a family is what makes the scaling statements
(\autoref{sec:scaling}) attributable to size rather than to family idiosyncrasy; the second and third
families guard the cross-family claims against Qwen3-specific idiosyncrasy. The oracle log-prob error
(\autoref{sec:framework}) references each model against a larger same-family oracle, chosen per
analysis. The gap-MSAR over-commitment fit (\autoref{tab:exp8b-msar}) uses Qwen3-14B over models
$0.6$--$8$B. The within-position cross-section (\autoref{fig:xsection}) and the variance-contraction intervention use
the family's largest model: Qwen3-32B over Qwen models ($0.6$--$14$B for the cross-section,
$4$--$14$B for the intervention) and OLMo-3.1-32B over OLMo-3-7B. The Qwen onset overlay and the onset
half-lives (\autoref{fig:onset-norm}, \autoref{tab:halflife}) share one re-score --- models
$0.6$--$8$B against the Qwen3-14B oracle --- and \autoref{app:oracle} repeats the onset analysis against
the Qwen3-32B oracle, finding the asymmetry unchanged. The position fit spans the full $0.6$--$32$B Qwen sweep. All
decoding is $\tau{=}1.0$,
$\mathrm{top\text{-}}p{=}1.0$ unless noted; the device is a single GPU (Modal L40S/H100 in the cloud,
MPS locally). The full project (inference-only, with no training or fine-tuning) fits within
roughly $25$ GPU-hours.

\begin{table}[htbp]
  \centering
  \small
  \begin{tabular}{lrl}
    \toprule
    \textbf{Model} & \textbf{Params} & \textbf{Role} \\
    \midrule
    \multicolumn{3}{l}{\emph{Qwen3} \citep{qwen3technical}} \\
    Qwen3-0.6B & 0.6B & model \\
    Qwen3-1.7B & 1.7B & model \\
    Qwen3-4B   & 4B   & model \\
    Qwen3-8B   & 8B   & model \\
    Qwen3-14B  & 14B  & model; oracle ($0.6$--$8$B) \\
    Qwen3-32B  & 32B  & model; oracle (onset, interv.) \\
    \midrule
    \multicolumn{3}{l}{\emph{OLMo-3} \citep{olmo3}} \\
    OLMo-3-7B-Instruct    & 7B  & model \\
    OLMo-3.1-32B-Instruct & 32B & oracle (OLMo) \\
    \midrule
    \multicolumn{3}{l}{\emph{Llama-3} \citep{llama3}} \\
    Llama-3.2-1B-Instruct & 1B & model \\
    Llama-3.2-3B-Instruct & 3B & model \\
    Llama-3.1-8B-Instruct & 8B & oracle (Llama) \\
    \bottomrule
  \end{tabular}
  \caption{One architecture per scale ladder isolates size; three independently-pretrained families
    guard against family idiosyncrasy. Within each family the oracle is the largest same-family model;
    Qwen3-14B is the oracle for the over-commitment analysis and Qwen3-32B for the within-position,
    onset, and intervention analyses. OLMo-3 and Llama-3 carry the onset-dynamics, bridge, and
  intervention results.}
  \label{tab:models}
\end{table}

\paragraph{Datasets.}
Three public, English benchmarks carry the analyses, one per measurement role.
\textbf{FActScore} biography prompts \citep{min2023factscore} ($64$ sampled topics)
supply long-form generations whose atomic facts are individually verifiable; these drive
the position fit and the teacher-forced decompositions.
\textbf{LongFact++} augmented prompts \citep{obeso2025realtime} ($401$ topics) supply the free-run
hallucinating trajectories the onset and intervention analyses run on.
\textbf{WildHallucinations} entities \citep{zhao2024wildhallucinations} ($20$ categories) --- referents from
real user--chatbot conversations --- carry the position fit into the wild: $48$ sampled with a fixed seed, balanced over
Wikipedia presence ($24$/$24$), spread across all $20$ categories, restricted to the middle perplexity band
(the extreme tails hallucinate at $\sim$$100\%$ or $\sim$$0\%$ at every rung and carry no positional signal).
All three are released for research use (FActScore MIT, LongFact Apache-2.0, WildHallucinations for research), as are the Qwen3
weights (Apache-2.0); our non-commercial academic use is consistent with each license, and
we collect and release no new data.

\paragraph{Verifier protocol.}
Claim labels come from a Claude verifier with web search. For a generated response
the verifier (i) extracts atomic factual claims, (ii) marks each claim's token span,
and (iii) labels each \emph{supported} (S) or \emph{not supported} (U) by issuing
web-search queries and checking the claim against retrieved evidence. The
\emph{fabrication onset} is the first token of the first U-labeled claim. For
\textbf{WildHallucinations} the referent is closed rather than web-wide: each claim is judged against
the dataset's curated per-entity corpus (the top-$10$ pages its authors scraped) by a tool-free Claude
call --- supported, contradicted-or-absent, or indecisive --- with no web search, using the same
claim-extraction leaf so granularity is identical across all three evals; indecisive labels are dropped, as
throughout. All
support/unsupport counts, snowball pairs, and intervention outcomes derive from these
labels; systematic verifier error would propagate, so we rely on cross-measurement
consistency (a verifier-independent white-box risk marker agrees with the labels at
AUROC $0.68$--$0.71$, \autoref{app:lowcommit}) rather than re-validating the verifier here. Double-judge
robustness and a verifier-temperature
sweep are open.

\paragraph{Reproducibility.}
Seeds are fixed at the outermost level and per-worker RNG is derived deterministically;
cloud runs use \texttt{aws s3 sync} for artifact persistence and resume to a
fresh-run-identical state. Every reported number is piped from one computation in the
corresponding analysis pipeline and regenerated from its source artifact, never hand-copied.
All confidence intervals are $95\%$ bootstrap intervals resampling the unit of analysis
(trajectories for the snowball and intervention results; claim-pairs for the
prefix-swap), and the source of variability is that resampling.

\section{Scaling and Knowledge Gaps}
\label{app:scaling}

The knowledge gap (intercept) falls monotonically with scale while knowledge degradation
(slope$\div$intercept) rises monotonically, over the underlying FActScore, facts-per-response,
abstention rate, and raw position slope (\autoref{tab:exp2-full}, expanding the two-sided trend of
\autoref{tab:scaling-twosided}). We report the degradation ratio because it normalizes the per-token
accumulation slope by the start-of-response gap.

\begin{table*}[tp]
  \centering
  \small
  \begin{tabular}{lrrrrrr}
    \toprule
    \textbf{Model} & \textbf{FActScore} & \textbf{facts/resp} & \textbf{frac no-fact} &
    \textbf{knowledge gap} & \textbf{raw slope} & \textbf{knowledge degradation} \\
    \midrule
    Qwen3-0.6B & 8.7  & 4.5 & 3.1\%  & 90.4\% & 2.7  & 3.0\%   \\
    Qwen3-1.7B & 13.4 & 6.4 & 0.0\%  & 82.5\% & 7.1  & 8.6\%   \\
    Qwen3-4B   & 28.1 & 7.3 & 1.6\%  & 56.6\% & 24.3 & 42.9\%  \\
    Qwen3-8B   & 47.5 & 7.1 & 12.5\% & 40.6\% & 17.7 & 43.5\%  \\
    Qwen3-14B  & 67.2 & 8.3 & 14.1\% & 21.7\% & 15.2 & 70.2\%  \\
    Qwen3-32B  & 65.1 & 7.9 & 10.9\% & 20.3\% & 22.3 & 110.1\% \\
    \bottomrule
  \end{tabular}
  \caption{The knowledge gap falls monotonically with scale while knowledge degradation rises
    monotonically --- the full fit behind \autoref{tab:scaling-twosided}. \emph{knowledge gap} is the
    OLS intercept (start-of-response hallucination rate) and \emph{knowledge degradation} is
  slope$\div$intercept, both in \%.}
  \label{tab:exp2-full}
\end{table*}

\paragraph{The knowledge-degradation metric is a mixed-effects fit.}
A linear-probability OLS slope$\div$intercept ratio divides by the start-of-response gap, so its
bootstrap CI explodes at the larger rungs where that intercept is small (the wide intervals the
two-sided fit would otherwise carry). \autoref{tab:scaling-twosided} therefore reports knowledge
degradation from a per-claim random-intercept logistic GLMM instead,
$\mathrm{logit}\,P(\text{supported}) = \beta_0 + \beta_1\,\text{position} + u_{\text{resp}}$, with a
random intercept per response absorbing topic/length baseline variation. The reported degradation is the
\emph{same} quantity as the OLS metric --- the relative rise in hallucination from the start
(position $0$) to the end (position $1$) of a response,
$[\,p_\text{hall}(1)-p_\text{hall}(0)\,]/p_\text{hall}(0)$ --- but read off the logistic fit, with a
$95\%$ credible interval from Monte-Carlo draws of $(\beta_0,\beta_1)$ under the mean-field variational
posterior; the result is the tight degradation CIs in \autoref{tab:scaling-twosided}.
\autoref{tab:exp2-glmm} reports the underlying fit: the per-position slope $\beta_{\text{pos}}$ is a
significant positional effect at nearly every model$\times$eval fit, so the degradation trend is genuine,
not an artifact of the ratio's shrinking denominator.

\begin{table}[t]
  \centering
  \small
  \setlength{\tabcolsep}{5pt}
  \begin{tabular}{l rr rr}
    \toprule
    & \multicolumn{2}{c}{\textbf{FActScore}} & \multicolumn{2}{c}{\textbf{LongFact++}} \\
    \cmidrule(lr){2-3} \cmidrule(lr){4-5}
    \textbf{Model} & $\beta_{\text{pos}}$ (95\% CI) & $\tau_{\text{resp}}$ & $\beta_{\text{pos}}$ (95\% CI) & $\tau_{\text{resp}}$ \\
    \midrule
    Qwen3-0.6B & $0.72\,[-0.15, 1.59]$ & $2.16$ & $0.40^{*}\,[0.27, 0.53]$ & $1.42$ \\
    Qwen3-1.7B & $0.82^{*}\,[0.26, 1.37]$ & $1.61$ & $0.75^{*}\,[0.65, 0.86]$ & $1.28$ \\
    Qwen3-4B & $1.78^{*}\,[1.35, 2.21]$ & $1.96$ & $1.00^{*}\,[0.90, 1.10]$ & $1.30$ \\
    Qwen3-8B & $1.12^{*}\,[0.70, 1.53]$ & $1.99$ & $0.98^{*}\,[0.88, 1.08]$ & $1.19$ \\
    Qwen3-14B & $1.09^{*}\,[0.69, 1.50]$ & $1.92$ & $0.81^{*}\,[0.71, 0.91]$ & $1.26$ \\
    Qwen3-32B & $1.65^{*}\,[1.24, 2.06]$ & $2.17$ & $1.14^{*}\,[1.00, 1.28]$ & $1.20$ \\
    \midrule
    OLMo-3-7B-Instruct & $1.17^{*}\,[0.72, 1.63]$ & $1.04$ & $0.54^{*}\,[0.39, 0.69]$ & $1.23$ \\
    OLMo-3.1-32B-Instruct & $0.32\,[-0.18, 0.82]$ & $3.01$ & $1.11^{*}\,[0.98, 1.24]$ & $1.09$ \\
    \midrule
    Llama-3.2-1B-Instruct & $1.89^{*}\,[1.42, 2.36]$ & $2.72$ & $0.54^{*}\,[0.42, 0.65]$ & $1.67$ \\
    Llama-3.2-3B-Instruct & $0.77^{*}\,[0.33, 1.20]$ & $2.22$ & $0.85^{*}\,[0.77, 0.94]$ & $1.30$ \\
    Llama-3.1-8B-Instruct & $1.23^{*}\,[0.83, 1.63]$ & $1.19$ & $0.90^{*}\,[0.83, 0.98]$ & $1.18$ \\
    \bottomrule
  \end{tabular}
  \caption{\textbf{Mixed-effects fit internals behind the knowledge-degradation column of \autoref{tab:scaling-twosided}.} Per Table-1 row and eval, the random-intercept logistic GLMM of atomic-fact support on relative within-response position: $\beta_{\text{pos}}$ is the per-position support log-odds slope (negated to read as degradation) with a Wald 95\% CI, and $\tau_{\text{resp}}$ the random-intercept SD (across-response baseline spread). The per-position effect is significant ($^{*}$, CI excludes $0$) at 20 of 22 model$\times$eval fits, so the degradation in \autoref{tab:scaling-twosided} is a genuine positional effect rather than an artifact of dividing by a shrinking start-of-response gap.}
  \label{tab:exp2-glmm}
\end{table}

\begin{figure*}[htbp]
  \centering
  \includegraphics[width=\textwidth]{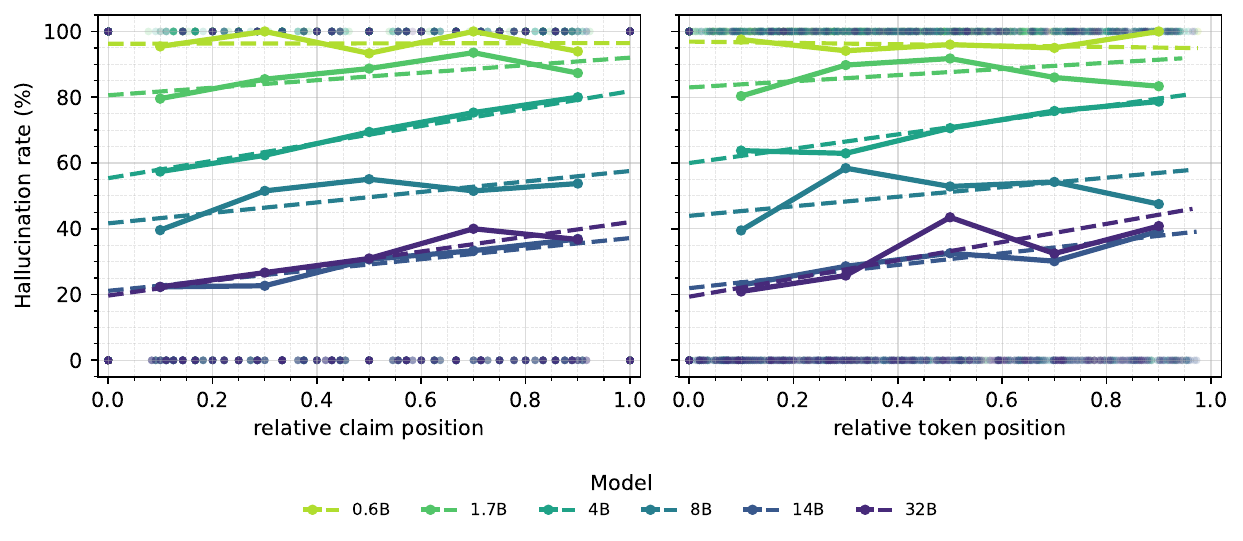}
  \caption{Hallucination rises roughly linearly with relative position, for both atomic facts
    (left) and tokens (right). Responses with $\ge5$ atomic facts, coarse-binned into five intervals
    (the first pools each response's opening facts); binning rather than smoothing avoids over-fitting
    the sparse near-zero region of the $\text{fact\_idx}/(n_\text{facts}{-}1)$ normalization. This
    per-response accumulation, its rate divided by the start-of-response gap, gives the knowledge-degradation
  metric of \autoref{tab:scaling-twosided}.}
  \label{fig:exp2-position}
\end{figure*}

\paragraph{Decoding ablation.}
The moment analyses are teacher-forced and so independent of decoding, but the free-run snowball is not. We
re-run it at greedy ($\tau{=}0$) and low-temperature ($\tau{=}0.5$) decoding on a shared $32$-prompt subset and
recompute the across-claim relative risk (\autoref{tab:decoding-ablation}). It stays above $1$ at every rung and
is larger at the biggest model than the smallest under all three policies, so the compounding is not an artifact
of sampling temperature; the small subset adds noise at the middle rungs.

\begin{table}[t]
  \centering
  \small
  \setlength{\tabcolsep}{4pt}
  \begin{tabular}{lrrrrr}
    \toprule
    \textbf{Decoding} & \textbf{Q0.6B} & \textbf{Q1.7B} & \textbf{Q4B} & \textbf{Q8B} & \textbf{Q14B} \\
    \midrule
    greedy ($\tau{=}0$) & $\times1.43$ & $\times2.05$ & $\times2.14$ & $\times1.66$ & $\times1.79$ \\
    $\tau{=}0.5$ & $\times1.22$ & $\times1.67$ & $\times1.85$ & $\times2.09$ & $\times1.97$ \\
    $\tau{=}1.0$ (canonical) & $\times1.07$ & $\times1.42$ & $\times1.57$ & $\times1.80$ & -- \\
    \bottomrule
  \end{tabular}
  \caption{\textbf{Reliability anti-scaling survives alternative decoding.} Across-claim snowball relative risk $\rr = P(U_{j+1}\mid U_j)/P(U_{j+1}\mid S_j)$ on mixed responses (\autoref{sec:mech:snowball}), per Qwen3 size, under greedy ($\tau{=}0$), low-temperature ($\tau{=}0.5$), and the canonical $\tau{=}1.0$ run, on a shared $32$-prompt subset. The relative risk stays above $1$ at every rung and is larger at the biggest model than the smallest under all three decoding policies (the small subset adds noise at the middle rungs), so the across-claim snowball is not an artifact of sampling temperature.}
  \label{tab:decoding-ablation}
\end{table}

\paragraph{WildHallucinations carries the scaling trend into the wild.}
The third block of \autoref{tab:scaling-twosided} fits the same per-claim random-intercept logistic GLMM
(\autoref{tab:exp2-glmm}) on the in-the-wild entities across all three family ladders (Qwen3 $0.6$--$32$B,
OLMo-3 $7$B/$32$B, Llama-3 $1$--$8$B), one draw per prompt at the standard decoding ($\tau{=}1.0$,
top-$p{=}1.0$, $512$ tokens); degradation is significant at $10$ of $11$ rungs. The trend survives a
knowledge-availability split: half the entities have no Wikipedia page --- the tail where a knowledge-gap
account would locate the failure. Fitting the two halves separately across all eleven rungs, degradation is
significantly positive at $10$ of $11$ on the Wikipedia half (to $+110$--$136\%$ at the largest rungs of
each family) and at $6$ of $11$ on the tail half, positive in sign at every rung. Knowledge availability
moves the gap; it does not remove the within-response decay.

\section{Onset Dynamics}
\label{app:onset-dyn}

\paragraph{The self-readable channel is impulsive; the oracle-referenced channels persist.}
Re-scoring the full next-token distribution around onset on the free-run trajectories, every
disagreement channel spikes together at onset, but only the oracle-referenced channels linger
(\autoref{tab:halflife}, \autoref{fig:onset-norm}): the self-readable $H(p_M)$ is impulsive, collapsing
with a half-life of $0.2$--$0.9$ tokens, while the oracle-referenced channels persist. The divergence
$\mathbb{E}[\delta]$ relaxes with half-life $2.5$--$3.1$ tokens, and the decoding risk
$\sqrt{\mathrm{Var}[\delta]}$ is the longest-lived ($3.4$--$4.6$ tokens), sustaining the highest
post-onset plateau of the three.

\begin{table*}[t]
  \centering
  \small
  \setlength{\tabcolsep}{4pt}
  \begin{tabular}{l rrrr rrrr rrrr}
    \toprule
    & \multicolumn{4}{c}{\textbf{$\mathbb{E}[\delta]$}} & \multicolumn{4}{c}{\textbf{$\sqrt{\mathrm{Var}[\delta]}$}} & \multicolumn{4}{c}{\textbf{$H(p_M)$}} \\
    \cmidrule(lr){2-5} \cmidrule(lr){6-9} \cmidrule(lr){10-13}
    \textbf{Model} & pre & onset & post & $t_{1/2}$ & pre & onset & post & $t_{1/2}$ & pre & onset & post & $t_{1/2}$ \\
    \midrule
    Qwen3-0.6B & 0.95 & 2.87 & 1.02 & 3.1 & 1.59 & 2.83 & 1.59 & 3.4 & 1.01 & 2.68 & 1.46 & 0.6 \\
    Qwen3-1.7B & 0.60 & 2.63 & 0.77 & 2.5 & 1.06 & 2.37 & 1.12 & 4.1 & 0.47 & 1.48 & 0.65 & 0.9 \\
    Qwen3-4B & 0.40 & 1.95 & 0.56 & 2.9 & 0.84 & 1.96 & 0.88 & 4.6 & 0.42 & 1.25 & 0.53 & 0.7 \\
    Qwen3-8B & 0.29 & 1.19 & 0.38 & 2.8 & 0.75 & 1.51 & 0.82 & 4.4 & 0.57 & 1.39 & 0.63 & 0.2 \\
    \bottomrule
  \end{tabular}
  \caption{\textbf{At a fabrication's onset the decoding risk is the longest-lived channel, the
    bias gap shorter, and the model's self-readable uncertainty the most impulsive.} The three
    \autoref{fig:onset-norm} channels of the disagreement variable (\autoref{eq:mse}) in nats --- the
    divergence $\mathbb{E}[\delta]{=}\mathrm{KL}(p_M\Vert p_O)$, the decoding std
    $\sqrt{\mathrm{Var}[\delta]}$, and the felt uncertainty $H(p_M)$ --- onset-aligned on free-running
    hallucinating trajectories (LongFact++; onset $t_0^\star$ = first unsupported atomic fact), the SAME
    frame \autoref{fig:onset-norm} plots. Models (Qwen3-0.6B--8B) scored against the Qwen3-14B oracle. Per channel the free-asymptote fit
    $C{+}A\,e^{-\lambda s}$ over the full post-onset window gives the pre-onset baseline
    ($t-t^\star{\in}[-5,-1]$), the onset peak, and the post-decay plateau $B{+}C$ (all nats), and the
    half-life $t_{1/2}{=}\ln 2/\lambda$ (tokens; $--$ where no clean exponential forms). The decoding std
    $\sqrt{\mathrm{Var}[\delta]}$ relaxes slowest ($3.4$--$4.6$ tokens) and sustains the highest
    plateau, the bias gap $\mathbb{E}[\delta]$ faster ($2.5$--$3.1$), and the self-readable $H(p_M)$
    the most impulsive ($0.2$--$0.9$).}  \label{tab:halflife}
\end{table*}

\section{Oracle Sensitivity}
\label{app:oracle}

\paragraph{The onset asymmetry survives a change of oracle size.}
$\delta$ is divergence from a particular oracle, so a fair worry is that the persistence asymmetry is an
idiosyncrasy of one reference model. It is not: re-scoring the same onset analysis against the Qwen3-14B
oracle (the cross-sectional reference) instead of the Qwen3-32B oracle leaves every qualitative invariant
intact (\autoref{tab:oracle-sensitivity}). Under both oracles the oracle-referenced channels --- divergence
$\mathbb{E}[\delta]$ and decoding risk $\sqrt{\mathrm{Var}[\delta]}$ --- stay elevated post-onset (the risk
half-life exceeds the realized-bias half-life $\overline{\delta(y_t)}^2$ at every rung), while the
self-readable $H(p_M)$ relaxes closest to baseline --- the same proxy break.
The two columns are scored on independently sampled free-run trajectory sets at different nucleus
truncations, so the exact half-lives differ and are not expected to match; that the \emph{asymmetry}
holds across oracle size on independent data, rather than on one shared sample, is the robustness claim.
A temperature-perturbed oracle would further extend this to oracle \emph{sharpness} (open,
\autoref{sec:limitations}); we carry out the cross-family extension to oracle \emph{family} next.

\begin{table*}[t]
  \centering
  \small
  \setlength{\tabcolsep}{5pt}
  \begin{tabular}{l rr rr rr rr rr}
    \toprule
    & \multicolumn{2}{c}{$\overline{\delta(y_t)}^2$ \textbf{$t_{1/2}$}} & \multicolumn{2}{c}{$\mathrm{Var}[\delta]$ \textbf{$t_{1/2}$}} & \multicolumn{2}{c}{$\mathrm{Var}[\delta]$ \textbf{po$/$pre}} & \multicolumn{2}{c}{$\mathbb{E}[\delta]$ \textbf{po$/$pre}} & \multicolumn{2}{c}{$H(p_M)$ \textbf{po$/$pre}} \\
    \cmidrule(lr){2-3}\cmidrule(lr){4-5}\cmidrule(lr){6-7}\cmidrule(lr){8-9}\cmidrule(lr){10-11}
    \textbf{Model} & 14B & 32B & 14B & 32B & 14B & 32B & 14B & 32B & 14B & 32B \\
    \midrule
    Qwen3-0.6B & 1.2 & 0.5 & 3.7 & 0.9 & $1.9\times$ & $1.7\times$ & $1.7\times$ & $1.6\times$ & $1.5\times$ & $1.3\times$ \\
    Qwen3-1.7B & 0.7 & 0.7 & 1.7 & 4.6 & $2.3\times$ & $1.8\times$ & $2.1\times$ & $1.5\times$ & $1.6\times$ & $1.2\times$ \\
    Qwen3-4B & 0.8 & 0.6 & 4.4 & 1.6 & $2.8\times$ & $1.8\times$ & $2.4\times$ & $1.8\times$ & $1.4\times$ & $1.3\times$ \\
    Qwen3-8B & 0.7 & 0.6 & 1.9 & 4.7 & $2.2\times$ & $2.6\times$ & $2.1\times$ & $1.9\times$ & $1.1\times$ & $1.2\times$ \\
    \bottomrule
  \end{tabular}
  \caption{\textbf{The onset asymmetry is invariant to oracle size.} The persistence asymmetry of
    \autoref{tab:halflife} --- the oracle-referenced channels (decoding risk $\mathrm{Var}[\delta]$,
    divergence $\mathbb{E}[\delta]$) stay elevated post-onset ($>1$) while the self-readable $H(p_M)$
    relaxes closest to baseline, the risk longest-lived --- reproduces whether Qwen3 models
    ($0.6$--$8$B) are scored against the Qwen3-14B or the Qwen3-32B oracle; the decoding-risk half-life
    $t_{1/2}$ exceeds the realized-bias half-life $\overline{\delta(y_t)}^2$ at every rung. po$/$pre is the
    post-onset ($t-t^\star\in[1,10]$) mean over the pre-onset baseline. \emph{The two columns are not a
    controlled swap}: they come from independently sampled free-run trajectory sets at different nucleus
    truncations (top-$k$ $1024$ vs $2048$), so the exact half-lives differ; the qualitative asymmetry does 
    not. Agreement across oracle size on independent data is, if anything, stronger than a shared-sample swap.}
  \label{tab:oracle-sensitivity}
\end{table*}

\begin{figure*}[t!]
  \centering
  \includegraphics[width=\textwidth]{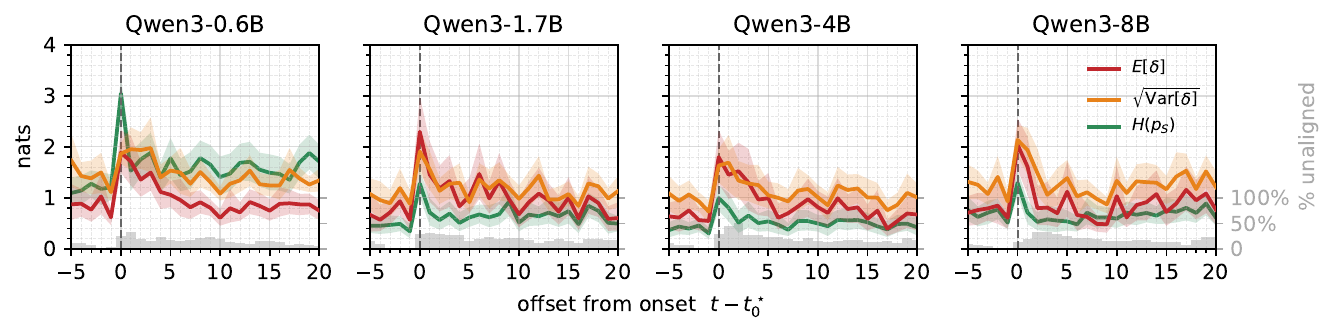}
  \caption{\textbf{The onset dynamics reproduce against a cross-family oracle: the self-readable channel again
    collapses within a token while the oracle-referenced channels persist.} The same onset-aligned free-run
    LongFact++ trajectories as \autoref{fig:onset-norm} (Qwen3 $0.6$--$8$B, one column per rung), re-scored
    against a distinct-family oracle (DeepSeek-V3-671B) via vocabulary-alignment transport: the three
    \autoref{fig:onset-norm} nats channels --- the bias gap $\mathbb{E}[\delta]{=}\mathrm{KL}(p_M\Vert p_O)$
    (red), the decoding risk $\sqrt{\mathrm{Var}[\delta]}$ (amber), and the self-readable $H(p_M)$ (green) ---
    as solid curves on one shared nats axis ($95\%$ CIs), the second row of \autoref{fig:onset-norm}. Per-channel
    half-lives and plateaus are in \autoref{tab:halflife-xfam}. The gray rug (secondary axis) is the \% of tokens
    not in a $1{:}1$ exact-aligned block at each offset (a branch-point approximation of the surface-bridged
    channels), the cross-family alignment-reliability context; positions are pre-filtered to $\ge\!80\%$ retained
    surface mass.}
  \label{fig:onset-xfam}
\end{figure*}

\paragraph{The onset asymmetry survives a change of oracle \emph{family}.}
Re-scoring the same onset trajectories against a distinct-family oracle (DeepSeek-V3-671B), aligned to the
Qwen3 vocabulary by prefix-routed transport, reproduces the ordering
$\sqrt{\mathrm{Var}[\delta]}{>}\mathbb{E}[\delta]{\gg}H(p_M)$ (\autoref{tab:halflife-xfam},
\autoref{fig:onset-xfam}): the self-readable $H(p_M)$ stays impulsive (half-life $0.4$ tokens), collapsing
while the oracle-referenced channels persist. The alignment injects per-position noise that blurs the fine
same-family separation --- $\mathbb{E}[\delta]$ and $\sqrt{\mathrm{Var}[\delta]}$ now relax with statistically
indistinguishable half-lives ($1.7$ vs $1.5$ tokens) --- so their ordering is carried by the sustained plateau
instead: $\sqrt{\mathrm{Var}[\delta]}$ settles at $1.19$ nats versus $\mathbb{E}[\delta]$'s $0.80$. Absolute
half-lives shorten against the more distant oracle, but the impulsive-versus-persistent proxy break is
unchanged.

\begin{table*}[t]
  \centering
  \small
  \setlength{\tabcolsep}{4pt}
  \begin{tabular}{l rrrr rrrr rrrr}
    \toprule
    & \multicolumn{4}{c}{\textbf{$\mathbb{E}[\delta]$}} & \multicolumn{4}{c}{\textbf{$\sqrt{\mathrm{Var}[\delta]}$}} & \multicolumn{4}{c}{\textbf{$H(p_M)$}} \\
    \cmidrule(lr){2-5} \cmidrule(lr){6-9} \cmidrule(lr){10-13}
    \textbf{Model} & pre & onset & post & $t_{1/2}$ & pre & onset & post & $t_{1/2}$ & pre & onset & post & $t_{1/2}$ \\
    \midrule
    Qwen3-0.6B & 0.83 & 1.89 & 0.83 & 2.1 & 1.43 & 1.87 & 1.43 & 2.0 & 1.18 & 3.05 & 1.60 & 0.0 \\
    Qwen3-1.7B & 0.69 & 2.30 & 0.90 & 1.2 & 1.00 & 1.91 & 1.12 & 0.9 & 0.44 & 1.28 & 0.63 & 0.3 \\
    Qwen3-4B & 0.62 & 1.80 & 0.66 & 2.7 & 0.95 & 1.63 & 1.00 & 2.2 & 0.39 & 1.00 & 0.47 & 1.1 \\
    Qwen3-8B & 0.74 & 2.04 & 0.82 & 0.8 & 1.20 & 2.14 & 1.22 & 1.0 & 0.67 & 1.29 & 0.67 & 0.2 \\
    \bottomrule
  \end{tabular}
  \caption{\textbf{The onset persistence ordering $\sqrt{\mathrm{Var}[\delta]}{>}\mathbb{E}[\delta]{\gg}H(p_M)$ survives a cross-family oracle.} The
    same onset-aligned LongFact++ trajectories as \autoref{tab:halflife} (Qwen3 $0.6$--$8$B; onset
    $t_0^\star$ = first unsupported atomic fact), re-scored against the cross-family DeepSeek-V3-671B oracle (a distinct model family from the Qwen3-0.6B--8B students). Per channel the free-asymptote
    fit $C{+}A\,e^{-\lambda s}$ gives the pre-onset baseline, onset peak, and post-decay plateau (nats),
    and the half-life $t_{1/2}$ (tokens). The self-readable $H(p_M)$ collapses fastest (half-life $0.4$
    tokens) while the oracle-referenced channels persist; among them $\sqrt{\mathrm{Var}[\delta]}$
    sustains the higher plateau ($1.19$ versus $0.80$ nats).}  \label{tab:halflife-xfam}
\end{table*}

\section{The Across-Claim Snowball}
\label{app:snowball}

The across-claim snowball of \autoref{sec:mech:snowball} survives the topic-difficulty control.
The relative risk $\text{RR}=P(U_{j+1}\mid U_j)/P(U_{j+1}\mid S_j)$ holds in
both samples. Across all responses the raw $\text{RR}_\text{all}$ climbs $\times1.28\to\times2.85$
with scale; restricting to \emph{mixed} responses (both a supported and an unsupported claim
present, so the topic is at least partly known --- controlling the base rate) attenuates it to
$\text{RR}_\text{mixed}$ $\times1.08\to\times1.71$ but does not remove it. Every $95\%$ CI ---
bootstrapped over responses, the independent unit --- excludes $1$ on thousands of consecutive
pairs per rung, so the snowball is neither a topic-gap confound nor a selection artifact of the
mixed restriction (\autoref{tab:exp8a}).

\begin{table*}[t]
  \centering
  \small
  \setlength{\tabcolsep}{5pt}
  \begin{tabular}{l r rr r@{\,$\pm$\,}l r rr r@{\,$\pm$\,}l}
    \toprule
    & \multicolumn{5}{c}{\textbf{all responses}} & \multicolumn{5}{c}{\textbf{mixed responses}} \\
    \cmidrule(lr){2-6} \cmidrule(lr){7-11}
    \textbf{Model} & $n$ & $\Pr(U{\mid}S)$ & $\Pr(U{\mid}U)$ & \multicolumn{2}{c}{RR} & $n$ & $\Pr(U{\mid}S)$ & $\Pr(U{\mid}U)$ & \multicolumn{2}{c}{RR} \\
    \midrule
    Qwen3-0.6B & $8{,}064$ & $0.74$ & $0.95$ & $1.28$ & ${\scriptstyle 0.07}$ & $2{,}416$ & $0.74$ & $0.80$ & $1.08$ & ${\scriptstyle 0.07}$ \\
    Qwen3-1.7B & $6{,}450$ & $0.44$ & $0.84$ & $1.93$ & ${\scriptstyle 0.12}$ & $3{,}877$ & $0.48$ & $0.68$ & $1.40$ & ${\scriptstyle 0.08}$ \\
    Qwen3-4B & $6{,}820$ & $0.33$ & $0.75$ & $2.25$ & ${\scriptstyle 0.15}$ & $4{,}972$ & $0.39$ & $0.60$ & $1.54$ & ${\scriptstyle 0.10}$ \\
    Qwen3-8B & $7{,}171$ & $0.23$ & $0.65$ & $2.85$ & ${\scriptstyle 0.22}$ & $5{,}361$ & $0.29$ & $0.50$ & $1.71$ & ${\scriptstyle 0.12}$ \\
    \bottomrule
  \end{tabular}
  \caption{\textbf{The across-claim snowball holds for both the raw and the topic-controlled sample --- not a confound or a selection artifact.} Consecutive claim pairs from Qwen3 free-run trajectories (\autoref{sec:mech:snowball}), shown for \emph{all} responses and the \emph{mixed} subset (holding both a supported and an unsupported claim, controlling the topic base rate). Per block: the consecutive-pair count $n$; the next claim's fabrication rate after a supported (spontaneous) vs.\ an unsupported claim, $\Pr(U{\mid}S)$/$\Pr(U{\mid}U)$; and the across-claim relative risk $\text{RR}{=}P(U_{j+1}{\mid}U_j)/P(U_{j+1}{\mid}S_j)$, $\pm$ half the $95\%$ CI bootstrapped over responses (the independent unit). RR clears $1$ in \emph{both} samples at every rung, so the lift is neither a topic-gap confound nor a selection artifact of the mixed restriction; controlled, $\text{RR}_\text{mixed}$ climbs $1.08\to1.71$ with scale.}
  \label{tab:exp8a}
\end{table*}

\section{The Variance-Contraction Intervention}
\label{app:contraction}

This section details the on-policy intervention behind \autoref{sec:mech:causal} and reports three results the
main text compresses: \emph{where} bias and variance concentrate (the live divergence event, not the verifier
label), how the contraction localizes to the inter-claim bridge, and a
dose-response that pins the lever to variance. The intervention runs on the LongFact++ free-run trajectories
(\autoref{app:setup}) for Qwen3 ($4$--$14$B, oracle $32$B) and OLMo-3-$7$B (oracle $32$B), $n{\approx}17$--$23$
source trajectories per model.

\paragraph{The mean-preserving contraction.}
At each step the co-resident oracle supplies the full next-token gap $\delta(x)=\log p_M(x)-\log p_O(x)$, its
mean $\mu=\mathbb{E}_{p_M}[\delta]=\mathrm{KL}(p_M\Vert p_O)$ (the \emph{bias}), and its spread
$\mathrm{Var}_{p_M}[\delta]$ (the \emph{decoding risk}); the intervention replaces $p_M$ with the contraction
$q_\lambda$ of \autoref{eq:contraction} (main text). The penalty acts on logits via $\delta$, so the knob is
invariant to a uniform shift of $\log p_M$ and reduces to ordinary temperature sharpening only in the
degenerate case $\delta\equiv\text{const}$; otherwise it is strictly a second-moment operation. The dose $\rho$
($\lambda$ chosen per step so $\mathrm{Var}_{q_\lambda}[\delta]=\rho\,\mathrm{Var}_{p_M}[\delta]$) is comparable
across models where a raw $\lambda$ is not. Holding the mean discards the \emph{lucky-good} excursions ($\delta$
on the supported side of $\mu$) symmetrically with the bad ones; removing only the bad ones would improve bias
and amount to leaking the oracle's answer, which \autoref{eq:contraction} deliberately does not do.

\paragraph{Solving $\nu$ and $\lambda$.}
Both multipliers are found by one-dimensional root-finding over the model nucleus support, exploiting that
each target moment is monotone in its knob. Inner solve: for a fixed penalty strength $\lambda$,
$\mathbb{E}_{q_\lambda}[\delta]$ is strictly increasing in $\nu$, so the unique $\nu$ with
$\mathbb{E}_{q_\lambda}[\delta]=\mu$ is bracketed by doubling outward from a spread-scaled guess and refined by
Brent's method (tolerance $10^{-10}$, $\le 200$ iterations); if the constraint already holds within $10^{-2}$
nats the tilt is skipped ($\nu{=}0$). Outer solve: $\mathrm{Var}_{q_\lambda}[\delta]$ is monotone decreasing
in $\lambda$ (with $\nu$ re-solved inside at each $\lambda$), so the $\lambda$ meeting the dose $\rho$ is
bracketed and Brent-refined the same way. At a degenerate position --- a near-collapsed support where a moment
is unreachable --- the solver caps at the bracket endpoint and records the realized
$\mathrm{var\_ratio}=\mathrm{Var}_{q_\lambda}[\delta]/\mathrm{Var}_{p_M}[\delta]$ and mean drift honestly
rather than aborting the co-resident generate barrier; the manipulation check
(\autoref{fig:exp9c-bias-hold}) reports these realized moments. Cost is a handful of Brent iterations of an
$O(|\text{nucleus}|)$ kernel per fired position, and the controller fires only at the $Z{=}2$ trigger
crossings (\emph{not} every token), so the per-token overhead is negligible against the model$+$oracle
forward pass it rides on.

\paragraph{Trigger and grid.}
Let $x^\star_t=\arg\max_x p_O(x)$ be the oracle's preferred token and $s_t=\log p_O(x^\star_t)-\log
p_M(x^\star_t)\ge 0$ the amount the model under-weights it. The controller fires where $s_t$ spikes
$Z$ standard deviations above its per-model mean ($s_t\ge\bar s+Z\sigma_s$, $Z{=}2$), standardized per model so
one threshold transfers across scales; anchoring at $x^\star$ (rather than $\arg\max_x\delta$, dominated by the
model's near-zero-probability tail) makes the signal ``oracle confident about a token the model misses,''
which pre-selects the recoverable knows-but-hallucinates regime. Around each trigger a tophat window of
$W{=}2$ tokens opens at offset $K$ and applies $q_\lambda$ at constant dose $\rho$. The swept grid is
$K\in\{-1,0,1,2\}\times\rho\in\{0.5,0.75,1.25,1.5\}$, with the controller refractory-gated so that it fires at
every crossing ($M{=}\infty$ in the arm tags). The control is no-op ($\lambda{=}0$,
the model's own CRN-identical free-run), giving one unambiguous paired baseline. We report the best arm over
the \emph{causal} $K{\ge}0$ subgrid only: a window opening before the online trigger ($K{=}{-}1$) cannot be
realized by any controller that detects the divergence before acting, so it is excluded even from the
oracle-in-the-loop ceiling. Generate is co-resident on a
single H100 (model$+$oracle, full vocabulary); verification is the same post-barrier, by-prompt
web-search pipeline as the observational tests (\autoref{app:setup}).

\paragraph{Bias and variance concentrate at the trigger, not the verifier onset.}
Per-position bias$^2$ and decoding risk are stored for every draw and plotted against two anchors: the
verifier fabrication onset $t_0^\star$ (first unsupported-claim token) and the live oracle trigger
$t_\mathrm{trig}$. The bias$^2$ excursion forms a sharp spike at offset $0$ in trigger coordinates but is
smeared and off-zero in onset coordinates: the verifier label \emph{lags} the model's computational
divergence by several tokens. Any intervention keyed to the verifier onset therefore aims downstream of where
bias and variance live, which is why relocating the anchor onto the live divergence event is what makes the
contraction land. One caveat bounds this: aligning on a threshold crossing mechanically shapes both moments
near offset $0$, so we report the ordering of the two spikes without testing it against a shuffled-trigger
null.

\paragraph{Localizing the contraction to the inter-claim bridge.}
The causal claim of \autoref{sec:mech:causal} is early$\to$late, so we relocate the same contraction onto the
bridge: the inter-claim gap $[c_{k^\star}.t_1,\,c_{k^\star+1}.t_0)$ of the first consecutive-unsupported claim
pair, fired at every gap token (structural --- no trigger, no threshold), with the single next claim
$c_{k^\star+1}$ as the outcome. The anchor is the verifier's claim labels and token spans on the control draw,
which the co-resident generate cannot know online, so the run is two-pass: pass one generates and verifies the
controls; pass two branches the arms from those spans at $\rho\in\{0.5,0.75\}$ and verifies. A draw qualifies on
a run of $\ge2$ consecutive unsupported claims plus $\ge1$ supported claim; intersecting the qualifying prompts
across a family's students yields per-family bridge sets --- $25$ of the $88$ pooled prompts for Qwen
($16$--$30$ control draws per student, fewer at scale as bigger models fabricate less), and comparably sized
sets for OLMo and Llama ($29$--$40$ control draws) --- with $15$--$31$ verifiable arm $c_{k^\star+1}$ per cell.
The verifier
force-searches every arm $c_{k^\star+1}$ and skips the arm's other claims --- the prefix $c_{\le k^\star}$ is
measured on the control, and claims past $c_{k^\star+1}$ are outside the outcome --- so no outcome claim is
dropped unassessed (\autoref{tab:exp9d-bridge}). Everything else --- kernel, common random numbers,
calibration, verification, and the trajectory-clustered paired bootstrap --- is the whole-response protocol
above, verbatim.

\begin{table}[htbp]
  \centering
  \small
  \setlength{\tabcolsep}{4pt}
  \begin{tabular}{lrrrrrc}
    \toprule
    \textbf{Model} & $\rho$ & $n$ & rate & $\Delta$ $[95\%\,\text{CI}]$ & S/NS/II & sig. \\
    \midrule
    Qwen3-1.7B & $0.50$ & 29 & 0.793 & $0.207$ $[0.069, 0.379]$ & 6/23/1 & $\bullet$ \\
    Qwen3-1.7B & $0.75$ & 29 & 0.724 & $0.276$ $[0.138, 0.448]$ & 8/21/1 & $\bullet$ \\
    Qwen3-4B & $0.50$ & 26 & 0.500 & $0.500$ $[0.308, 0.692]$ & 13/13/3 & $\bullet$ \\
    Qwen3-4B & $0.75$ & 27 & 0.481 & $0.519$ $[0.333, 0.704]$ & 14/13/2 & $\bullet$ \\
    Qwen3-8B & $0.50$ & 18 & 0.667 & $0.333$ $[0.111, 0.556]$ & 6/12/1 & $\bullet$ \\
    Qwen3-8B & $0.75$ & 18 & 0.611 & $0.389$ $[0.167, 0.611]$ & 7/11/1 & $\bullet$ \\
    Qwen3-14B & $0.50$ & 15 & 0.533 & $0.467$ $[0.200, 0.733]$ & 7/8/1 & $\bullet$ \\
    Qwen3-14B & $0.75$ & 15 & 0.400 & $0.600$ $[0.333, 0.800]$ & 9/6/1 & $\bullet$ \\
    Olmo-3-7B & $0.50$ & 22 & 0.636 & $0.364$ $[0.182, 0.545]$ & 8/14/4 & $\bullet$ \\
    Olmo-3-7B & $0.75$ & 21 & 0.524 & $0.476$ $[0.286, 0.668]$ & 10/11/6 & $\bullet$ \\
    Llama-3.2-1B & $0.50$ & 31 & 0.742 & $0.258$ $[0.129, 0.419]$ & 8/23/4 & $\bullet$ \\
    Llama-3.2-1B & $0.75$ & 28 & 0.750 & $0.250$ $[0.107, 0.393]$ & 7/21/9 & $\bullet$ \\
    Llama-3.2-3B & $0.50$ & 27 & 0.630 & $0.370$ $[0.185, 0.556]$ & 10/17/12 & $\bullet$ \\
    Llama-3.2-3B & $0.75$ & 31 & 0.613 & $0.387$ $[0.226, 0.548]$ & 12/19/9 & $\bullet$ \\
    \bottomrule
  \end{tabular}
  \caption{\textbf{Contracting the inter-claim bridge prevents the second fabrication (absolute effects).} Per student and dose $\rho$: the verifiable ($n$) count and web-verified unsupported rate of the second fabrication's slot $c_{k^\star+1}$, the within-trajectory paired $\Delta$ (control $-$ arm) with its trajectory-clustered $95\%$ bootstrap CI, and the arm's Supported/Not-Supported/Insufficient-Information label counts. The control rate is $1.0$ by construction --- the bridge is the inter-claim gap of a \emph{consecutive-unsupported pair}, so the control's second claim is unsupported --- and $\Delta$ is therefore the fraction of second fabrications the contraction prevents. A bullet ($\bullet$) marks a CI excluding zero: $\Delta = 0.21$--$0.60$ at all fourteen student$\times$dose cells ($n = 15$--$31$). Every arm $c_{k^\star+1}$ is force-searched by the verifier, so no claim is dropped unassessed (\autoref{app:contraction}). As in \autoref{tab:exp9c-causal}, the co-resident oracle makes this the \emph{ceiling} of an online inter-claim controller, not a deployable decoding rule.}
  \label{tab:exp9d-bridge}
\end{table}

\paragraph{Dose-response: contraction helps, expansion does not.}
At $M{=}\infty,K{=}0$, the paired $\Delta$ is positive for every contraction dose ($\rho{<}1$) at all six rungs.
On the four Qwen3 and OLMo rungs it is near-zero or negative for every expansion dose ($\rho{>}1$); this
contraction-vs-expansion gap (deeper variance suppression helps, variance inflation does not) is the robust
signal that the lever is variance. The $\rho{=}1.25$ and $\rho{=}1.5$ cells are identical because the kernel
clamps the expansion side; on the two Llama-3.2 rungs that clamp lands on a contraction, so their nominal
expansion cells also reduce the rate ($\Delta\approx0.21$/$0.26$) and provide no clean expansion control. The
ordering \emph{within} the contraction side ($\rho{=}0.5$ vs $0.75$) is model-specific. This is the
dose-response image of \autoref{eq:contraction} acting on the variance term of \autoref{eq:mse} at fixed
$\mathrm{KL}$.

\paragraph{Absolute effects.}
\autoref{tab:exp9c-causal} carries the absolute rates behind the relative reduction of
\autoref{fig:indecision-b}: the best-arm paired $\Delta$ is $0.13$--$0.33$ with a $95\%$ bootstrap CI excluding
zero at all six model$\times$family rungs, and the winning arm always sits on the contraction side
($\rho{\le}0.75$). The manipulation check (\autoref{fig:exp9c-bias-hold}) confirms the realized variance falls to
the dose while the per-fired-step bias drift stays $\sim$11 orders of magnitude below $|\mu|$, so the reduction
is attributable to the decoding-risk channel and not to a residual mean shift.

\begin{table}[htbp]
  \centering
  \small
  \setlength{\tabcolsep}{4pt}
  \begin{tabular}{lrrrrrc}
    \toprule
    \textbf{Model} & ctrl. & best & $\Delta$ $[95\%\,\text{CI}]$ & $K$ & $\rho$ & sig. \\
    \midrule
    Qwen3-4B & 0.374 & 0.219 & $0.158$ $[0.020, 0.313]$ & $+1$ & $0.5$ & $\bullet$ \\
    Qwen3-8B & 0.346 & 0.128 & $0.221$ $[0.131, 0.312]$ & $+2$ & $0.5$ & $\bullet$ \\
    Qwen3-14B & 0.219 & 0.088 & $0.131$ $[0.054, 0.202]$ & $0$ & $0.5$ & $\bullet$ \\
    OLMo-3-7B-Instruct & 0.606 & 0.394 & $0.211$ $[0.041, 0.391]$ & $+1$ & $0.5$ & $\bullet$ \\
    Llama-3.2-1B & 0.544 & 0.216 & $0.247$ $[0.125, 0.360]$ & $+1$ & $0.5$ & $\bullet$ \\
    Llama-3.2-3B & 0.457 & 0.119 & $0.332$ $[0.261, 0.399]$ & $+1$ & $0.5$ & $\bullet$ \\
    \bottomrule
  \end{tabular}
  \caption{\textbf{Contracting decoding risk causally lowers downstream hallucination (absolute effects).} Per rung: the no-op control and best-contraction-arm web-verified unsupported-claim rates over the rest of the response, the within-trajectory paired $\Delta$ (control $-$ best) with its trajectory-clustered $95\%$ bootstrap CI, and the winning arm's onset offset $K$ and dose $\rho$. A bullet ($\bullet$) marks a CI that excludes zero: $\Delta = 0.13$--$0.33$ at all six model$\times$family rungs. The best arm is selected post hoc per rung over the causal ($K\ge0$) grid, so $\Delta$ is the oracle-in-the-loop \emph{ceiling} of an online risk controller, not a deployable decoding rule. The relative reduction is the page-1 teaser (\autoref{fig:indecision-a}).}
  \label{tab:exp9c-causal}
\end{table}

\paragraph{No length, claim-count, or informativeness confound.}
The contraction changes \emph{which} claims a response supports, not how many it makes or how long it runs.
\autoref{tab:exp9c-confounds} holds each rung's best arm against control on four descriptive statistics: length
is pinned by the generation cap (within $0.5\%$), fixing the rate's denominator; the atomic-claim count --- the
informativeness signal at fixed length --- moves non-directionally within $-9.7\%$ to $+4.2\%$; and self-BLEU
(stylistic diversity, normalized to $[0,1]$) shifts a mean $+1.0\%$, up on three rungs and down on three.
Only the firing rate moves
decisively (zero for control). A sub-$10\%$ change in claim count at fixed length cannot produce a $35$--$74\%$
drop in the unsupported \emph{fraction}: the effect is orthogonal to, and far larger than, any descriptive
shift.

\begin{table}[htbp]
  \centering
  \small
  \setlength{\tabcolsep}{4pt}
  \begin{tabular}{llrrrrr}
    \toprule
    \textbf{Model} & arm & $n$ & tokens & claims & self-BLEU & fires \\
    \midrule
    Llama-3.2-1B & control & 86 & 254 & 6.69 & 0.103 & 0.0 \\
     & $K{=}+1,\ \rho{=}0.5$ & 86 & 253 & 6.17 & 0.098 & 47.9 \\
    \midrule
    Llama-3.2-3B & control & 84 & 256 & 8.30 & 0.138 & 0.0 \\
     & $K{=}+1,\ \rho{=}0.5$ & 84 & 255 & 7.91 & 0.139 & 29.4 \\
    \midrule
    OLMo-3-7B & control & 22 & 256 & 7.68 & 0.097 & 0.0 \\
     & $K{=}+1,\ \rho{=}0.5$ & 22 & 256 & 7.32 & 0.091 & 29.7 \\
    \midrule
    Qwen3-4B & control & 21 & 256 & 8.52 & 0.118 & 0.0 \\
     & $K{=}+1,\ \rho{=}0.5$ & 21 & 256 & 8.19 & 0.108 & 19.9 \\
    \midrule
    Qwen3-8B & control & 23 & 256 & 8.48 & 0.144 & 0.0 \\
     & $K{=}+2,\ \rho{=}0.5$ & 23 & 256 & 7.65 & 0.151 & 16.0 \\
    \midrule
    Qwen3-14B & control & 20 & 256 & 8.25 & 0.106 & 0.0 \\
     & $K{=}0,\ \rho{=}0.5$ & 20 & 256 & 8.60 & 0.127 & 15.5 \\
    \bottomrule
  \end{tabular}
  \caption{\textbf{The hallucination reduction is not an artifact of shorter or sparser responses.} Per rung, the no-op control and the best-contraction arm (the same $K\ge0$, $M{=}\infty$ selection as \autoref{tab:exp9c-causal}) on three descriptive statistics of the free-run response: length (\emph{tokens}, fixed by the generation cap), emitted atomic-claim count (\emph{claims} --- at fixed length, also the informativeness signal), and stylistic diversity (\emph{self-BLEU}; higher $=$ less diverse). The best arm shifts the emitted-claim count by at most $10\%$ and self-BLEU by at most $0.021$ relative to control --- neither distinguishable from no intervention --- while the controller \emph{fires} on $16$--$48$ tokens per draw (\emph{fires}, $0$ for control). A model that fabricated less by emitting fewer or shorter claims would register here; none does, so the $35$--$74\%$ reduction (\autoref{tab:exp9c-causal}) is a smaller unsupported \emph{fraction} at fixed, equally informative output, attributable to the contracted decoding-risk channel.}
  \label{tab:exp9c-confounds}
\end{table}

\section{Detector Implementation}
\label{app:detectors}

\paragraph{Semantic entropy.}
The detector we test is a functional of $p_M$ alone --- the object \autoref{pred:blindness} concerns. The
blindness test (\autoref{sec:mech:blindness}, \autoref{tab:detector-blindness}) scores each claim with
the original semantic-entropy detector \citep{kuhn2023semantic,farquhar2024detecting}: the off-shelf
\texttt{deberta-large-mnli} natural-language-inference model, bidirectional-entailment clustering of resamples,
and the discrete cluster-proportion entropy (nats). The resamples are the $K{=}25$ same-model stochastic draws
of each prompt already produced for the free-run trajectories (\autoref{app:setup}). One adaptation makes it
claim-level: for the claim under test, each resample's NLI stance toward the claim (premise $=$ resample,
hypothesis $=$ claim) $\in\{\text{entailment, neutral, contradiction}\}$ is its semantic cluster, and the score
is the discrete entropy over those clusters; the cluster count is therefore not a free hyperparameter (emergent
from entailment, $\le3$ at the claim level). Higher entropy $=$ more resample disagreement about the claim $=$
a stronger hallucination flag. SelfCheck-Prompt \citep{manakul2023selfcheckgpt}, also a $p_M$-functional over
the same draws, is implemented from its canonical leave-one-out consistency prompt.

\section{Commitment Regimes Give a Label-Free Risk Marker}
\label{app:lowcommit}

\paragraph{Three commitment regimes.}
A per-model three-state model over the per-token triple (bias $\mathbb{E}[\delta]$,
entropy $H(p_M)$, decoding risk $\sqrt{\mathrm{Var}[\delta]}$) recovers the regimes
of \autoref{fig:token-risk}: \emph{grounded} (low bias, entropy \& risk), \emph{precarious} (low
  bias \& entropy but high risk, a confident yet high-variance self-conditioned branch; $\mathrm{prec}$
below), and \emph{diverged} (high bias, entropy \& risk) --- gap regimes, not correctness. The
trajectory runs grounded, shifts to diverged at the verifier onset, then persists in the precarious
state through the continuation --- the self-conditioning signature (conditioning on its own draw) ---
and the gap-only fit below shows it recurs at every scale.

\paragraph{A gap-only fit recovers the same over-commitment signature.}
A three-state Markov-switching fit to the gap $\delta_t=\log p_M-\log p_O$ alone is
BIC-preferred over two states for $99.9$--$100\%$ of series at every scale. Its highest state is
the over-commitment regime: highest mean gap and also highest variance (an over-commitment, not a
low-variance ``confidently-wrong'' attractor), where the model
over-rates the realized token relative to the oracle (\autoref{tab:exp8b-msar}). It is reachable
from the precarious state and sticky (entry and self-transition probabilities both
$\approx0.42$--$0.45$ at every scale), and its over-rating magnitude shrinks monotonically with
scale, $2.4$ nats at $0.6$B to $0.8$ nats at $8$B, matching the falling bias$^2$ share of
\autoref{fig:xsection}.

\begin{table}[htbp]
  \centering
  \small
  \setlength{\tabcolsep}{4pt}
  \begin{tabular}{lrrrrr}
    \toprule
    & \textbf{prefers} & \multicolumn{2}{c}{\textbf{over-commit state}} & \multicolumn{2}{c}{\textbf{occupancy}} \\
    \cmidrule(lr){3-4}\cmidrule(lr){5-6}
    \textbf{Model} & $k{=}3$ & $\mu_\text{oc}$ & $\sigma_\text{oc}$ & U / S & AUROC \\
    \midrule
    Qwen3-0.6B & 99.9\% & 2.36 & 2.75 & 0.42 / 0.25 & 0.681 \\
    Qwen3-1.7B & 100\%  & 1.73 & 2.35 & 0.46 / 0.27 & 0.710 \\
    Qwen3-4B   & 100\%  & 1.26 & 1.98 & 0.43 / 0.25 & 0.713 \\
    Qwen3-8B   & 100\%  & 0.81 & 1.56 & 0.44 / 0.26 & 0.714 \\
    \bottomrule
  \end{tabular}
  \caption{Unsupported tokens load the over-commitment state $\approx1.7\times$ as often as supported
    ones, separating the two at AUROC $0.68$--$0.71$ with no verifier and no labels. Three-state
    gap-MSAR at the BIC-selected $k{=}3$ on free-run trajectories (Qwen3 $0.6$--$8$B models, $14$B
    oracle); $k{=}3$ is preferred over $k{=}2$ for $99.9$--$100\%$ of series. The over-commitment state
    (highest mean gap $\mu_\text{oc}$) is also the highest-variance ($\sigma_\text{oc}$); its over-rating
  magnitude shrinks with scale ($2.36\to0.81$ nats). AUROC is stable across all four scales.}
  \label{tab:exp8b-msar}
\end{table}

\paragraph{Hallucinated tokens load the over-commitment state.}
This backs \autoref{sec:scaling:s3}: unsupported atomic-fact tokens occupy the over-commitment
state $\approx1.7\times$ as often as supported ones ($0.42$--$0.46$ vs $0.25$--$0.27$),
separating supported from unsupported content at AUROC $0.68$--$0.71$, \emph{stable across all
four scales} (\autoref{tab:exp8b-msar}). The state is a label-free, white-box risk marker (read off
the logits with no verifier), and it agrees with the independent FActScore position fit
(\autoref{app:scaling}) about where in a response the model is most likely to fabricate.

\paragraph{The precarious state bridges adjacent fabrications.}
The bridge effect reads the \emph{precarious} state of the three-channel fit above, not the gap-only
over-commitment state of \autoref{tab:exp8b-msar}: between consecutive claims, the confident precarious state
fills the inter-claim bridge more when the next claim is also unsupported (\autoref{sec:mech:bridge},
\autoref{fig:bridge}). \autoref{tab:bridge} gives the per-rung effect at the
$\le10$-token adjacency window: the probability a bridge token is in the \emph{precarious} state when the next
claim is a fabrication $\Pr(\mathrm{prec}\mid\mathrm{U}\to\mathrm{U})$ vs.\ supported
$\Pr(\mathrm{prec}\mid\mathrm{U}\to\mathrm{S})$, their absolute difference, and the ratio
$\Pr(\mathrm{prec}\mid\mathrm{U}\to\mathrm{U})/\Pr(\mathrm{prec}\mid\mathrm{U}\to\mathrm{S})-1$, alongside the
number of prev-fabrication pairs $n$ behind each estimate. The effect is scale-emergent: strongest at the
largest ($14$B) rung, with Qwen3-1.7B the lone negative rung. Rungs with fewer than $100$ prev-fabrication
pairs (the small Llama-3.2 rungs) are too sparse to estimate the ratio reliably and are excluded from
\autoref{fig:bridge}.

\begin{table*}[htbp]
  \centering
  \small
  \setlength{\tabcolsep}{6pt}
  \begin{tabular}{lrrrrr}
    \toprule
    \textbf{Model} & $n$ & $\Pr(\mathrm{prec}\mid \mathrm{U}\to\mathrm{S})$ & $\Pr(\mathrm{prec}\mid \mathrm{U}\to\mathrm{U})$ & $\Delta$ & $\frac{\Pr(\mathrm{prec}\mid \mathrm{U}\to\mathrm{U})}{\Pr(\mathrm{prec}\mid \mathrm{U}\to\mathrm{S})}-1$ \\
    \midrule
    Qwen3-0.6B$^{\dagger}$ & 58 & 0.54 & 0.63 & $+0.09$ & $+17\%$ \\
    Llama-3.2-1B-Instruct$^{\dagger}$ & 4 & 0.33 & 1.00 & $+0.67$ & $+200\%$ \\
    Qwen3-1.7B & 130 & 0.47 & 0.45 & $-0.02$ & $-4\%$ \\
    Llama-3.2-3B-Instruct$^{\dagger}$ & 9 & 0.40 & 0.76 & $+0.35$ & $+89\%$ \\
    Qwen3-4B & 163 & 0.36 & 0.38 & $+0.02$ & $+5\%$ \\
    OLMo-3-7B-Instruct & 102 & 0.19 & 0.21 & $+0.02$ & $+12\%$ \\
    Qwen3-8B & 170 & 0.38 & 0.44 & $+0.06$ & $+15\%$ \\
    Qwen3-14B & 178 & 0.23 & 0.40 & $+0.16$ & $+69\%$ \\
    \bottomrule
  \end{tabular}
  \caption{\textbf{Per-rung values behind the precarious-regime bridge effect (\autoref{fig:bridge}).}
    Every model rung (all families, size-ordered): the number of prev-fabrication claim pairs
    $n$, the probability that an inter-claim bridge token is in the \emph{precarious} state when the next
    claim is supported $\Pr(\mathrm{prec}\mid \mathrm{U}\to\mathrm{S})$ vs.\ a fabrication $\Pr(\mathrm{prec}\mid \mathrm{U}\to\mathrm{U})$, their absolute difference $\Delta = \Pr(\mathrm{prec}\mid \mathrm{U}\to\mathrm{U}) - \Pr(\mathrm{prec}\mid \mathrm{U}\to\mathrm{S})$,
    and the plotted ratio $\Pr(\mathrm{prec}\mid \mathrm{U}\to\mathrm{U})/\Pr(\mathrm{prec}\mid \mathrm{U}\to\mathrm{S})-1$ (the \autoref{fig:bridge} $y$-axis). Adjacency window
    $\le 10$ tokens, mixed-trajectory restricted (as \autoref{fig:bridge}).
    $^{\dagger}$~Daggered rungs are excluded from \autoref{fig:bridge}: outside the $1$--$14$\,B scale window (Qwen3-0.6B) or fewer than 100 prev-fabrication pairs (Llama-3.2-1B-Instruct, Llama-3.2-3B-Instruct).}
  \label{tab:bridge}
\end{table*}

\end{document}